\documentclass[table]{article} 
\usepackage{collas2024_conference,times}
\usepackage{easyReview}

\usepackage{amsmath,amsfonts,bm}

\def\eqref#1{equation~\ref{#1}}

\def\floor#1{\lfloor #1 \rfloor}
\def\1{\bm{1}}

\DeclareMathAlphabet{\mathsfit}{\encodingdefault}{\sfdefault}{m}{sl}
\SetMathAlphabet{\mathsfit}{bold}{\encodingdefault}{\sfdefault}{bx}{n}

\usepackage{hyperref}
\hypersetup{
    colorlinks=true,
    linkcolor=red,
    filecolor=magenta,
    urlcolor=blue,
    citecolor=purple,
    pdftitle={Overleaf Example},
    pdfpagemode=FullScreen,
    }
\usepackage[inline]{enumitem}
\usepackage{url}
\usepackage{graphicx}
\usepackage{subfig}
\usepackage{booktabs}
\usepackage{multirow}
\usepackage{placeins}
\usepackage{shortbold}
\usepackage[english]{babel}
\newtheorem{theorem}{Theorem}

\newcommand{\Cg}{\cellcolor{lightgray}}

\title{Chunking: Continual Learning is not just about Distribution Shift}

\author{Thomas L.~Lee \& Amos Storkey \\
School of Informatics \\
University of Edinbrugh \\
\texttt{\{T.L.Lee-1@sms.,A.storkey@\}ed.ac.uk}
}

\collasfinalcopy 

\begin{document}

\maketitle

\begin{abstract}
Work on continual learning (CL) has thus far largely focused on the problems arising from shifts in the data distribution. However, CL can be decomposed into two sub-problems: (a) shifts in the data distribution, and (b) dealing with the fact that the data is split into chunks and so only a part of the data is available to be trained on at any point in time. In this work, we look at the latter sub-problem, the \emph{chunking} of data. We show that chunking is an important part of CL, accounting for around half of the performance drop from offline learning in our experiments. Furthermore, our results reveal that current CL algorithms do not address the chunking sub-problem, only performing as well as plain SGD training when there is no shift in the data distribution. Therefore, we show that chunking is both an important and currently unaddressed sub-problem and until it is addressed CL methods will be capped in performance. Additionally, we analyse why performance drops when learning occurs on identically distributed chunks of data, and find that forgetting, which is often seen to be a problem due to distribution shift, still arises and is a significant problem. We also show that performance on the chunking sub-problem can be increased and that this performance transfers to the full CL setting, where there is distribution shift. Hence, we argue that work on chunking can help advance CL in general.\footnote{Code is available at \href{https://github.com/Tlee43/Chunking-Setting}{https://github.com/Tlee43/Chunking-Setting}} 
\end{abstract}

\section{Introduction}
How should we update a neural network efficiently when we observe new data? This issue remains an open problem, and is one that the field of \emph{continual learning} (CL) addresses. Many methods \citep{Delange2021A, Parisi2019review, wang2023comprehensive} and settings \citep{hsu2018re, antoniou2020defining, van2019three} have been proposed in recent years. Specifically, CL studies settings where a learner sees a stream of chunks of data and where the data distribution for each chunk changes over time. This type of change in the data distribution is commonly known as \emph{task shift} \citep{caccia2020online}. 

CL can be decomposed into two sub-problems: (a) learning with a changing data distribution, and (b) only having access to a single chunk of data for learning at any point in time, unable to ever re-access previous chunks. We call this latter sub-problem the \emph{chunking problem}. Current work in CL has focused on realistic settings where both sub-problems are present and where their separation has not been made explicit. This has meant that separating the two sub-problems contributions to the difficulty of CL and to what extent CL methods address each sub-problem has been largely unexplored. Therefore, in this work we investigate the chunking problem, looking at the extent to which it is a factor of CL performance and how well current CL methods deal with it. To do this we formulate and look at the \emph{chunking setting} where we remove the task-shift element of CL but keep everything else the same. We do not propose this setting to be like the real world, where there is often task-shift, but to analyse the chunking sub-problem and its contribution to the more realistic full CL setting with task shift. 

Our analysis of the chunking setting establishes a number of findings. First, we show that chunking is responsible for a significant part of the performance difference between CL and offline learning---learning with full access to all the data. Second, our experiments demonstrate that current CL methods do not address the chunking sub-problem, performing comparably to plain SGD training in the chunking setting. These two points suggest that chunking is a significant and unaddressed problem in CL. Additionally, we demonstrate that a large amount of forgetting---the loss of knowledge learnt from previously seen data---occurs in the chunking setting. This casts doubt on the common sentiment that forgetting is caused mainly by task shift \citep{lee2021continual, ramasesh2020anatomy}. Last, we demonstrate that we can reduce forgetting and improve performance in the chunking setting, using a per-chunk weight averaging scheme. This performance improvement transfers to the full CL setting—--where there is also task shift—--establishing that work on the chunking sub-problem has the potential to impact CL in general. 

The main contributions of this work are:
\begin{itemize}
    \item Formulation of the \emph{chunking sub-problem} of CL and demonstrating that it is the reason for a large part of the performance drop between offline learning and CL.
    \item Analysis of chunking, where we show among other things that current CL methods do not address this sub-problem, performing similarly to plain SGD training.
    \item Demonstrating that performance in the chunking setting can be improved and that this performance transfers to the full CL setting, illustrating how work on chunking can help improve CL in general. 
\end{itemize}
\begin{figure*}[t]
\begin{center}
\includegraphics[trim={0cm 9.7cm 0cm 4.3cm}, scale=0.6, clip]{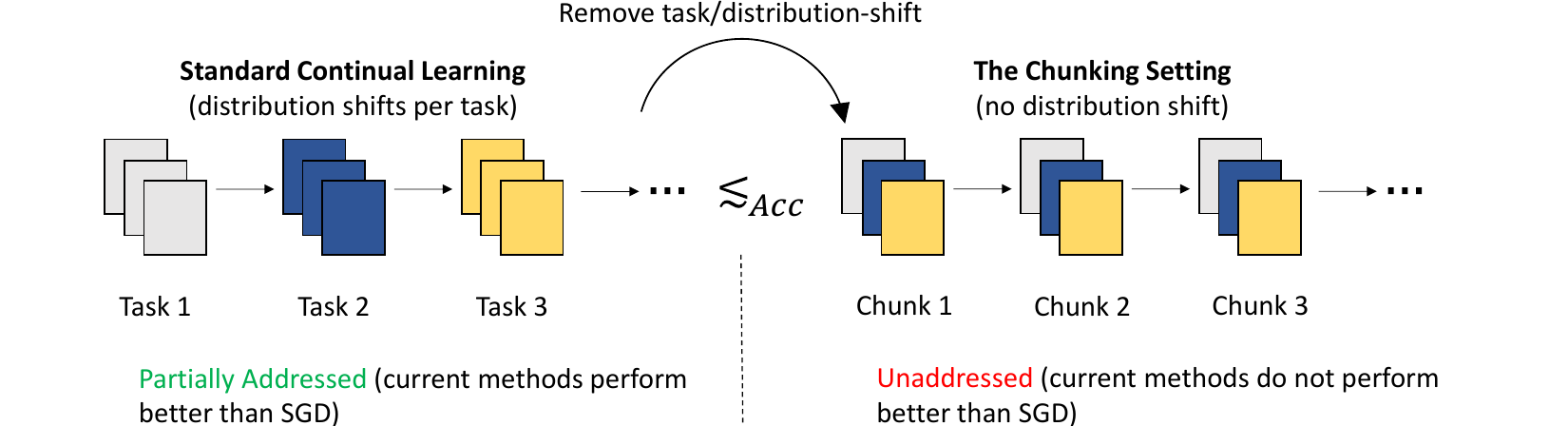}
\end{center}
\caption{\textbf{Standard continual learning versus the chunking setting}. In standard continual learning (CL) a learner sequentially receives chunks of data called tasks and there is a shift in distribution between each task. While in the chunking setting, each chunk of data is identically distributed. CL methods have become better at dealing with task-shift and so partially address standard continual learning. However, we show that current CL methods do not tackle the chunking sub-problem as they perform no better than plain SGD training in the chunking setting. We also find that chunking contributes a significant part of the performance gap between offline learning and CL. This means that performance in the chunking setting is significantly lower than offline learning performance, for commonly used numbers of chunks. Therefore, as the chunking setting provides an approximate upperbound to performance in CL, improving capability in the chunking setting is a necessity to obtain high-performing CL methods.} 
\label{fig:chunking}
\end{figure*}

\section{Preliminaries and Related Work}
\label{sec:prelim}
Continual learning (CL) is a well-studied problem, with many different settings and methods being proposed \citep{van2019three, wu2022pretrained, Mirzadeh2020Understanding, Delange2021A}. We focus on classification problems. In this context, the standard CL setting (sometimes called offline CL \citep{prabhu2020gdumb}), consists of a learner seeing a sequence of tasks. Each \emph{task} consists of a single chunk of data, where the data is drawn i.i.d. from a task-specific data distribution. In practice, this most commonly means that each task consists of all the training data from a subset of classes in the dataset \citep{van2019three}. A learner only views each task once and can only revisit data from previous tasks which it has stored in a limited memory buffer. For example, for CIFAR-10 \citep{krizhevsky2009learning} a learner might first see all the data for the airplane and ship classes, then see the data from the dog and cat classes and so on, seeing all data from two classes at a time until the learner has seen all the classes. Standard CL is further subdivided into different scenarios depending on the type of distribution shift which occurs and if the learner is told when the task changes or not. Four popular CL scenarios which are all refinements of the standard CL setting are task-incremental, class-incremental, domain-incremental and task-free learning \citep{van2019three, aljundi2018task, wang2023comprehensive}. Importantly, as these settings contain task-shift they are not the same as the chunking setting we look at, which is a sub-problem of all of them. Additionally, our results show the behaviour of CL methods in the chunking setting is quite different to their behaviour in each of the above standard CL scenarios. This is because, as shown in Figure~\ref{fig:chunking}, CL methods improve quite a bit upon plain SGD training in standard CL but perform similarly to it in the chunking setting. This suggest that current CL methods are quite good at dealing with the task-shift sub-problem but do not currently tackle the chunking problem, a key insight of our work.       

The chunking sub-problem of CL is closely related to online learning, without task shift \citep{Hoi202Online, bottou2003large}. In both cases the data is observed in the form of a stationary data stream. However, in chunking the data is batched into chunks to match modern neural network learning processes. Straight online learning can be seen as a special case when each chunk consists of one data instance. Furthermore, we investigate the neural network case in contrast to much work in online learning which focuses on the linear case \citep{Hoi202Online}. There is recent work on online learning of neural networks, for example \citet{ash2020warm, caccia2022anytime}; and \citet{sahoo2017online}. But, their focus is not on linking or comparing their work to CL and often the settings and assumptions are quite different from CL. Another related line of work to this paper is the study of the \emph{new-instance} setting \citep{lomonaco2017core50, prabhu2023computationally}. However, usually in this setting there is task shift \citep{lomonaco2017core50} and if not, to the best of our knowledge, previous work has not focused on how the setting is related to CL---with task shift. This is unlike this paper which focuses on providing insight into CL and where the chunking setting is deliberately constructed to examine the chunking sub-problem of CL. Also, in addition to the related work discussed in this section, we also discuss how chunking relates to positive transfer in Appendix~\ref{appen:posTransfer} and it use in analysing the online CL setting in Appendix~\ref{appen:onlineCL}. More generally, there are many areas in machine learning, like federated learning \citep{zhang2021survey}, which have challenges similar to that of the chunking problem. Therefore, these areas might also benefit from work looking at chunking and vice versa. 

The reason why the chunking sub-problem and its relation to CL has not been thoroughly explored in the CL literature is unclear. We see it as perhaps due to the fact that while the continual learning of neural networks has a long history \citep{deAgulo95, polikar01, storkey1997increasing, grossberg88}, the current focus has been on realistic settings that contain a large amount of task shift \citep{Delange2021A, wang2023comprehensive}. Because of this greater complexity, most work has focused on reducing the negative effects of task shift, leaving the component of performance due to chunking to be overlooked and implicit. Yet decomposing a problem can aid its solution, and indeed, we show that chunking is responsible for a large part of the performance drop from offline learning to CL. 

To see if it is possible to improve performance in the chunking setting we consider per-chunk weight averaging and find that it provides significant improvements. There have been many weight averaging approaches proposed for offline learning \citep{izmailov2018averaging, tarvainen2017mean}. Additionally, there have been weight averaging methods suggested for CL \citep{lee2017overcoming, lee2020residual, garg2023tic, stojanovski2022momentum}. So, we are not proposing our per-chunk weight averaging scheme as being a particularly original method. Instead, its use here is new and of interest in that it demonstrates that we can improve performance in the chunking setting and that this performance transfers to the full CL setting with task shift. To the best of our knowledge this has not been demonstrated before.

\section{The Chunking Setting}
In the \emph{chunking setting}, a learner sees a sequence of chunks $C_1, C_2, \ldots, C_N$, and trains on one chunk of data at a time, where chunks are not revisited. Each chunk of data consists of instance pairs $(x,y)$, with $x \in X$ (e.g. images) and labels $y \in Y$. The data in all chunks are drawn from the same distribution, so there is no distribution shift. Furthermore, in this paper, to control for class imbalance effects we consider a \emph{balanced} chunking setting (henceforth assumed); we constrain each chunk to have as close to the same number of instances for each class in $Y$ as possible. In this way we ensure the results of our experiments are solely due to the effects of limited data availability through chunking and not due to class imbalance. We record results for the case where the chunks are class imbalanced in Appendix~\ref{Appen:Sampling} and observe that for our experimental setup class imbalance does not have any significant effect. 

In practice, to perform experiments in the chunking setting, consider a class-balanced training dataset of size $M$ and a target chunk size $S$. First, we randomly reorder the training data for each class, and then arrange all the data into a class-ordered list. Data is then sequentially allocated into $\floor{M/S}$ chunks by assigning each element of the list in turn to a chunk, in a cyclical fashion. So, the first data item goes into chunk 1, second into chunk 2 etc., up to an item into chunk $\floor{M/S}$, then the next into chunk 1 again and so on. Then we randomly permute the data within each chunk and randomly reorder the chunks themselves. To ensure chunks are fully balanced, in the experiments in this paper we choose chunk sizes so that all chunks are of equal size and contain the same number of data instances for each class. Finally, we reserve a equal-sized portion of data from each class to form a test set which is used to evaluate the accuracy of a method. 

The only difference between the chunking setting and the full CL setting is the lack of task shift. Therefore, the chunking setting provides a simple way to analyse and understand the problems caused by chunking. Importantly, we are \emph{not} proposing the chunking setting as being \emph{realistic}, instead we use it to explore the chunking sub-problem which is a component of all the more realistic settings where there is also task-shift. Also, performance in the chunking setting gives an approximate upper bound to CL performance (task-shift just makes things harder), and so without solving this setting CL will never be able to improve beyond current chunking performance. 
 
\section{Analysis of the Chunking Setting}
\label{sec:Analysis}
To see how much chunking is a factor of the performance in CL, we perform an experiment with DER++ \citep{buzzega2020dark}, a popular and highly capable CL method. We find that chunking plays a significant part in the performance drop from the offline setting. The experiment consists of comparing the relative performance drop from offline SGD training to DER++ on the standard CL and chunking settings. For the experiment, we use class-incremental learning for standard CL; which means at test time, the learner has to classify across all the classes \citep{van2019three} in exactly the same way as the chunking setting. Additionally, following the experimental setup of \citet{buzzega2020dark} and \citet{boschini2022class}, we use a ResNet18 backbone and a 10 task/chunk split of \begin{enumerate*}[label=(\alph*)]
\item CIFAR-100 \citep{krizhevsky2009learning} with a memory size of 2000, and
\item Tiny ImageNet \citep{Stanford2015Tiny} with a memory size of 5120. \end{enumerate*} The rest of the experimental details are given in Appendix~\ref{appen:expDetails}. The results are presented in Table~\ref{tab:compToCL} and show that the performance drop between offline learning and chunking is $50.05\%$ and $46.69\%$ of the full performance drop from offline learning to CL for CIFAR-100 and Tiny ImageNet, respectively. This indicates that a significant part of the performance drop of CL from offline learning is due to chunking and not due to task/distribution shift. Also, in the real world it is often the case that the hard task shifts commonly used in continual learning do not happen \citep{Bang2021Rainbow, bang2022online, mi2020generalized} and instead there are smoother changes between tasks which could reduce the effect of task shift and increase the importance of dealing with chunking.
\begin{table*}[t]
  \caption{Accuracy of DER++ when using a ResNet18 in the offline, chunking and standard CL class-incremental settings, along with the percentage drop in accuracy from offline learning to CL due to chunking (Chunking Prop.). We split each dataset into 10 tasks following the experimental setup of \citet{buzzega2020dark} and \citet{boschini2022class}. The table shows, by the bold column, that a significant proportion of the performance drop from offline learning to CL is due to the chunking problem.}
  \label{tab:compToCL}
  \centering 
  \begin{tabular}{llllll}
    \toprule
    Dataset & Offline  & Chunking  & CL & \textbf{Chunking Prop.} \\
    \midrule
    CIFAR-100  & $73.72_{\pm 0.115}$ & $63.35_{\pm 0.348}$& $53.00_{\pm 0.327}$ &  \boldmath$50.05\%$ \\
    Tiny ImageNet  & $60.63_{\pm0.366}$ & $50.54_{\pm0.118}$ & $39.02_{\pm0.97}$ &  \boldmath$46.69\%$ \\
    \bottomrule
  \end{tabular}
\end{table*}

\subsection{Performance in the Chunking setting}
\begin{figure}[t]
    \centering
    \subfloat{{\includegraphics[trim={0 0 0 0mm}, clip, width=.5\linewidth]{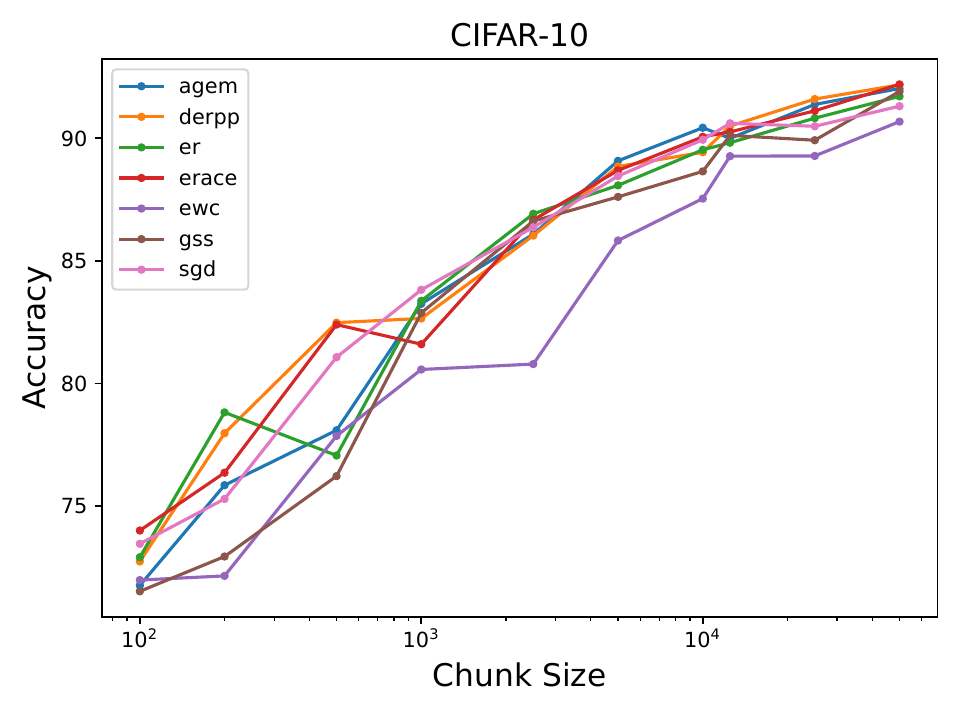} }}%
    \subfloat{{\includegraphics[trim={0 0 0 0mm}, clip, width=.5\linewidth]{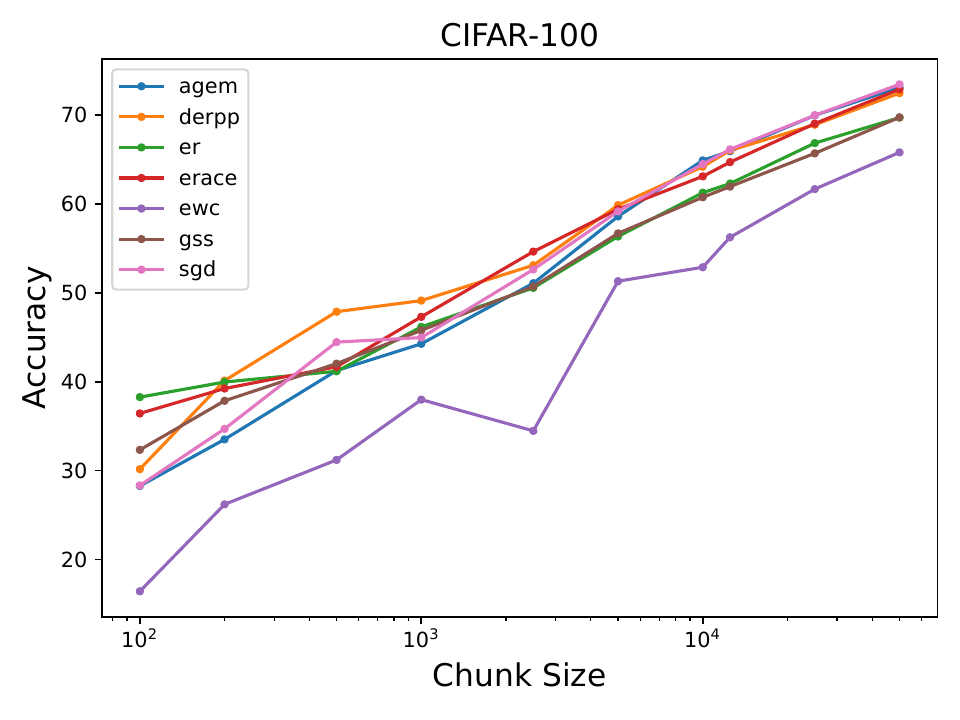} }}%
    \caption{End-of-training accuracy against chunk size on CIFAR-10 and CIFAR-100. Each data point on a curve presents the end-of-training accuracy of a method from a full run with chunks of the size given on the horizontal axis. The plots show that a smaller chunk size leads to a greater performance drop from offline learning (the performance of the right most point in each plot) and that CL methods perform similarly to plain SGD training in the chunking setting.}
    \label{fig:NNChunkingCurves}
\end{figure}
\begin{figure}[t]
    \centering
    \begin{minipage}{0.49\textwidth}
        \centering
        \includegraphics[trim={0 0 0 0mm}, clip, width=1\textwidth]{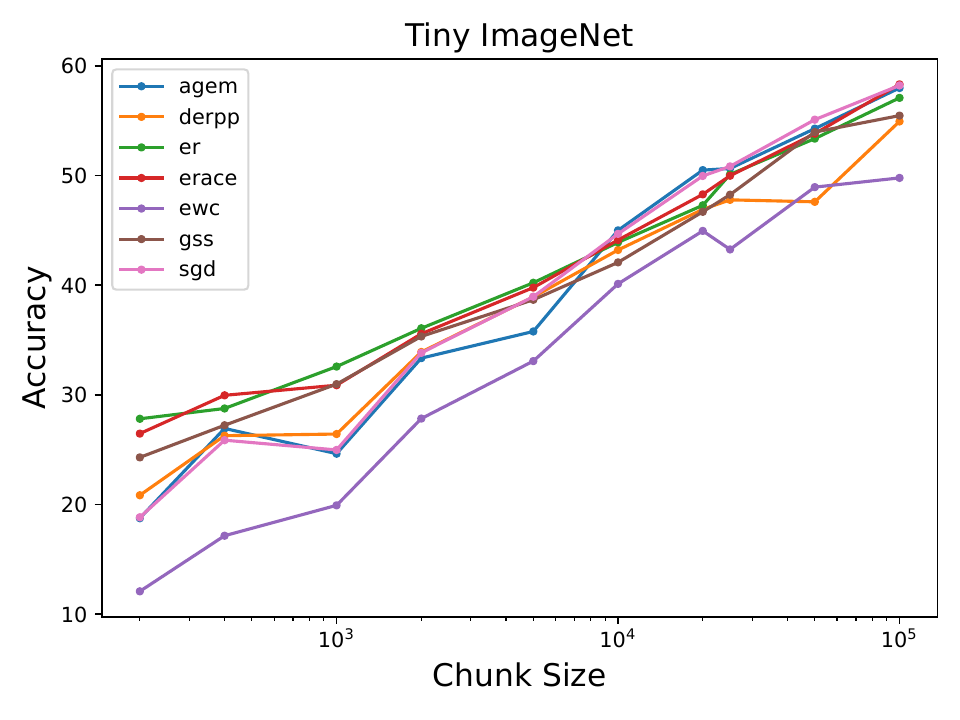} 
        \caption{End-of-training accuracy against chunk size for Tiny ImageNet. Each data point on a curve presents the end-of-training accuracy of a method from a full run on Tiny ImageNet with chunks of the size given on the horizontal axis. The plot shows that the smaller the chunk size the greater the performance drop from offline learning (the performance of the right most point in the plot) and that CL methods perform similarly to plain SGD training  in the chunking setting.}
        \label{fig:tinyimgChunkingCurve}
    \end{minipage}\hfill
    \begin{minipage}{0.49\textwidth}
        \centering
        \vspace{-11.5mm}
        \includegraphics[trim={0 0 0 0mm}, clip, width=1\textwidth]{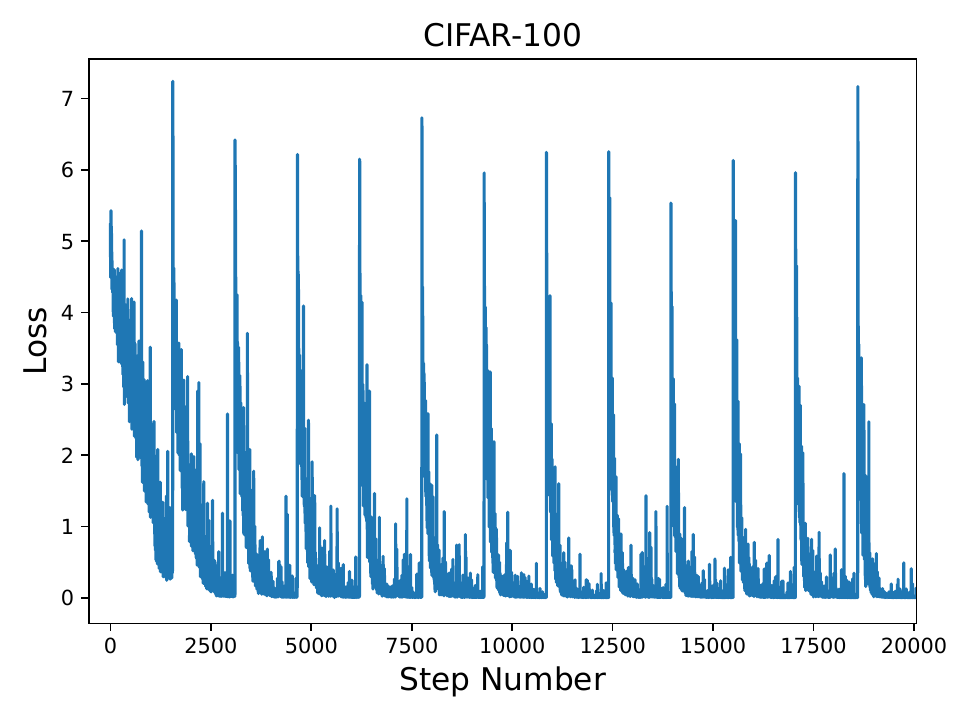} 
        \caption{The training loss curve for plain SGD on CIFAR-100 when training on 50 chunks, where we plot the training loss for the first 2000 update steps corresponding to learning on the first 13 chunks. The plot shows that the loss converges for each chunk and hence that the learner does not underfit when training on any chunk.}
        \label{fig:50ChunkLossCurve}
    \end{minipage}\hfill
    \vspace{-5mm}
\end{figure}
Our results on the chunking setting show that CL methods perform no better than plain SGD training. For instance, Figures~\ref{fig:NNChunkingCurves} and \ref{fig:tinyimgChunkingCurve} present the performance of state-of-the-art CL methods for different chunk sizes and a memory buffer size of 500 examples on CIFAR-10, CIFAR-100 and Tiny Imagenet, which are commonly used in the CL literature \citep{Delange2021A}. We train on each chunk for 50 epochs for CIFAR-10 and CIFAR-100 and 100 epochs for Tiny ImageNet which we found to be give the best or comparable to the best performance (as shown in Appendix~\ref{appen:epochNum}). The full experimental details of this experiment are described in Appendix~\ref{appen:expDetails}. The results displayed in Figures~\ref{fig:NNChunkingCurves} and \ref{fig:tinyimgChunkingCurve} show that all the CL methods perform roughly the same as plain SGD training. Hence, our results indicate that current CL methods do not tackle the chunking problem and instead have focused on reducing the performance drop due to task shift, as they perform much better than SGD on settings with task shift \citep{wang2023comprehensive}. One point to note on this is that the replay methods ER \citep{Chaudhry2020Continual} and ER-ACE \citep{caccia2021new} perform better than SGD for very small chunk sizes. This is due to them storing 500 examples in memory and so have an effective chunk size of 500 more data points than SGD, which impacts performance for chunk sizes around and below 500. Additionally, Figures~\ref{fig:NNChunkingCurves} and \ref{fig:tinyimgChunkingCurve} show that there is a large performance drop as the chunk size decreases. For example, on CIFAR-100 for offline learning when all the data is in one chunk, corresponding to a chunk size of 50000, CL methods get a test accuracy of around $73\%$ but when each chunk consists of 1000 examples they get around $45\%$. Also, in addition to the results here, we show that the stability gap phenomenon \citep{de2023continual} appears in the chunking setting in Appendix~\ref{Appen:StabilityGap} and look at the impact of using pretrained models in the chunking setting in Appendix~\ref{appen:pretrain}.

\begin{figure}[t]
    \centering
    \subfloat{{\includegraphics[trim={0 0 0 0mm}, clip, width=.5\linewidth]{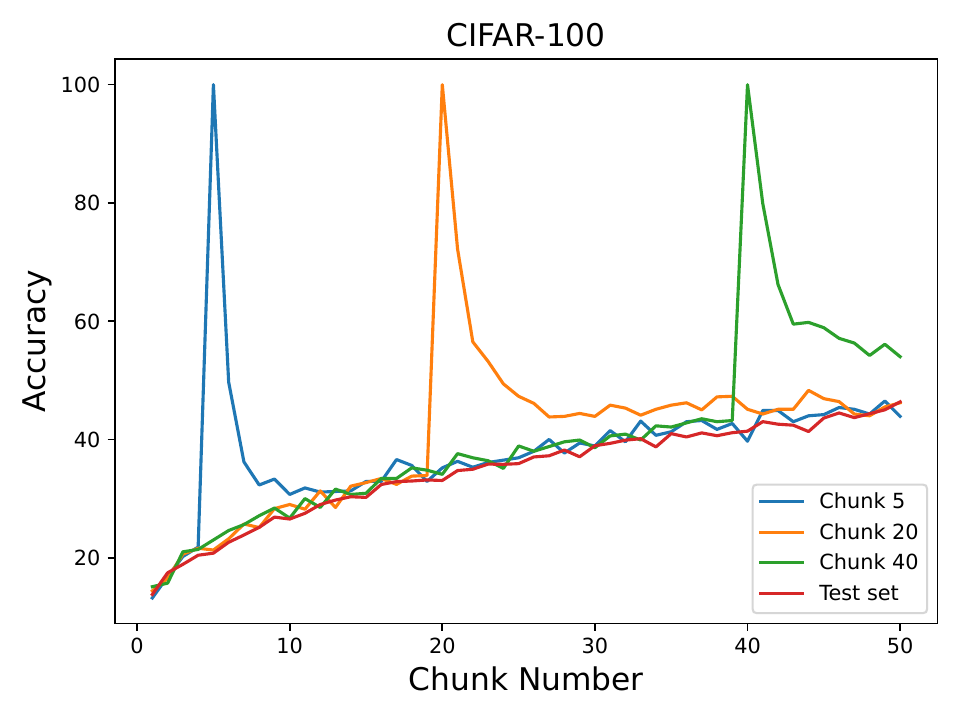} }}%
    \subfloat{{\includegraphics[trim={0 0 0 0mm}, clip, width=.5\linewidth]{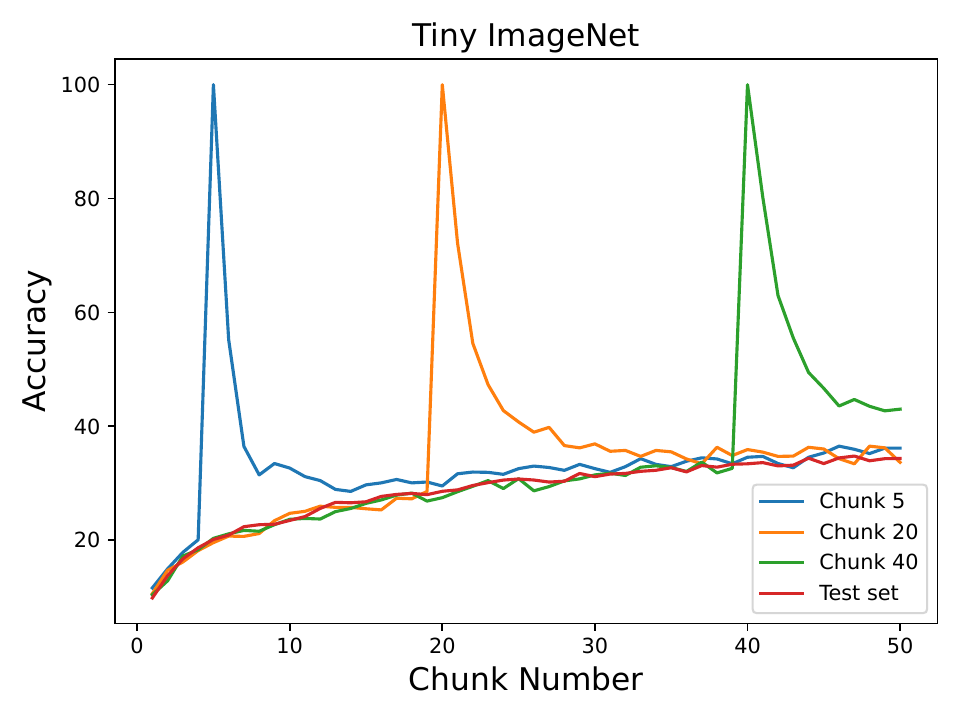} }}%
    \caption{Accuracy at the end of learning on each chunk for the training set of the $5^{th}$, $20^{th}$ and $40^{th}$ chunks and the test set, for CIFAR-100 and Tiny ImageNet when using plain SGD training. We split the datasets into 50 chunks, corresponding to a chunk size of 1000 and 2000 for CIFAR-100 and Tiny ImageNet, respectively. The plots show that after learning on a chunk the accuracy on that chunk quickly drops to the level of test set performance and hence that the learner quickly forgets a large part of the knowledge of a chunk after learning on it.}
    \label{fig:50ChunkForgettingCurve}
\end{figure}
An important question to ask is why does chunking reduce the performance from offline learning. There are three general possibilities: not integrating all the information a chunk has into the model (underfitting), fully integrating each chunk's information but at the cost of forgetting previous information (forgetting) or a mixture of both. To explore which possibility is true we look in more detail at the case when we train using 50 chunks. We present the training loss curve in Figure~\ref{fig:50ChunkLossCurve} of learning on the first 13 chunks for CIFAR-100 and in Figure~\ref{fig:50ChunkForgettingCurve} the test accuracy and accuracy on the training data for the $5^{th}$, $20^{th}$ and $40^{th}$ chunks evaluated at the end of each chunk, for CIFAR-100 and Tiny ImageNet (see Appendix~\ref{appen:add50ChunkForgettingCurve} for CIFAR-10). The training loss curve in Figure~\ref{fig:50ChunkLossCurve} shows that we fit each chunk well as the loss plateaus for each chunk and at a relatively low value. Furthermore, the accuracy curves for each chunks training data in Figure~\ref{fig:50ChunkForgettingCurve} establishes that after training on the chunk the model fits it perfectly, achieving an accuracy of $100\%$. Hence, we know that the learner fits each chunk well, removing the possibility of underfitting. Figure~\ref{fig:50ChunkForgettingCurve} also shows that after learning on the chunk the accuracy on that chunks data quickly drops back to the level of the test set performance, showing that the learner is forgetting a lot of the chunk's information. Therefore, our results suggest that the performance drop in the chunking setting is due to forgetting. However, not all of a chunk's information is forgotten as the test accuracy improves as the learner sees more chunks. Additionally, our results indicate that forgetting is not only due to task shift, which is commonly assumed in previous work \citep{lee2021continual, ramasesh2020anatomy}, but that it is also due to chunking. It is also useful to question why forgetting is occurring. One potential reason is overfitting, as Figure~\ref{fig:50ChunkForgettingCurve} shows that the learner achieves $100\%$ accuracy on the current chunk. However, whether this means overfitting is happening is a difficult question to answer as a model can achieve $100\%$ training accuracy and still generalise well---as demonstrated by the double decent phenomenon \citep{nakkiran2021deep}. If overfitting is the cause of the forgetting, it is interesting that even at the start of learning the learner forgets previous chunks. This is because, at the start of learning there is plenty of capacity in the network to both overfit to the current chunk and retain knowledge of previous chunks. So, if this is the case, the current way models are learnt seems to be particularly destructive of previous knowledge. 

Our results suggest that forgetting is the reason for the reduced performance in the chunking setting, when compared to offline learning. However, not all the information provided by a chunk is forgotten and if a learner could repeatedly resample chunks it would approach offline performance (as suggested by Figure~\ref{fig:50ChunkForgettingCurve}). This fact is standard knowledge for online learning \citep{bottou2003large} and has recently been shown to be true for CL with task-shift \citep{lesort2023challenging}. However, unlike standard online learning, due to real-world constraints in the chunking setting and CL it is not possible to resample chunks. Therefore, in these settings we need to be able to fully learn a chunk of data without needing to repeatedly revisit it in the future. This implies that improving chunking performance and reducing forgetting is closely related to improving the efficiency of learning. Hence, we hope that work on improving chunking performance will also improve the general efficiency of learning algorithms and vice versa.    

\section{Per-Chunk Weight Averaging}
\begin{figure}[t]
    \centering
    \subfloat[]{{\includegraphics[trim={0 0 0 0mm}, clip, width=.5\linewidth]{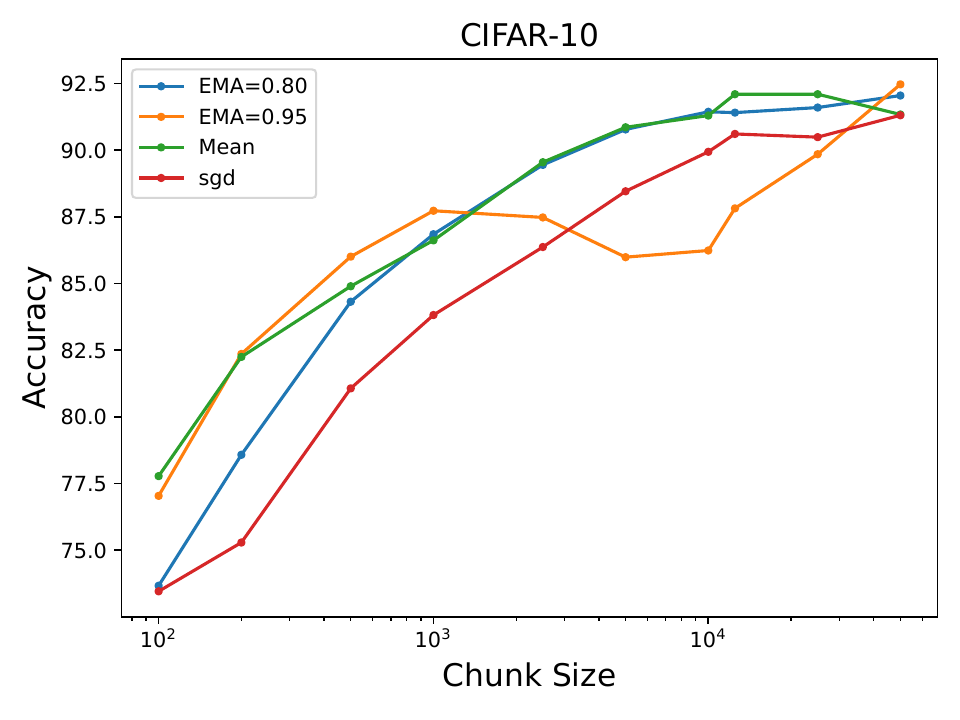} }}%
    \subfloat[]{{\includegraphics[trim={0 0 0 0mm}, clip, width=.5\linewidth]{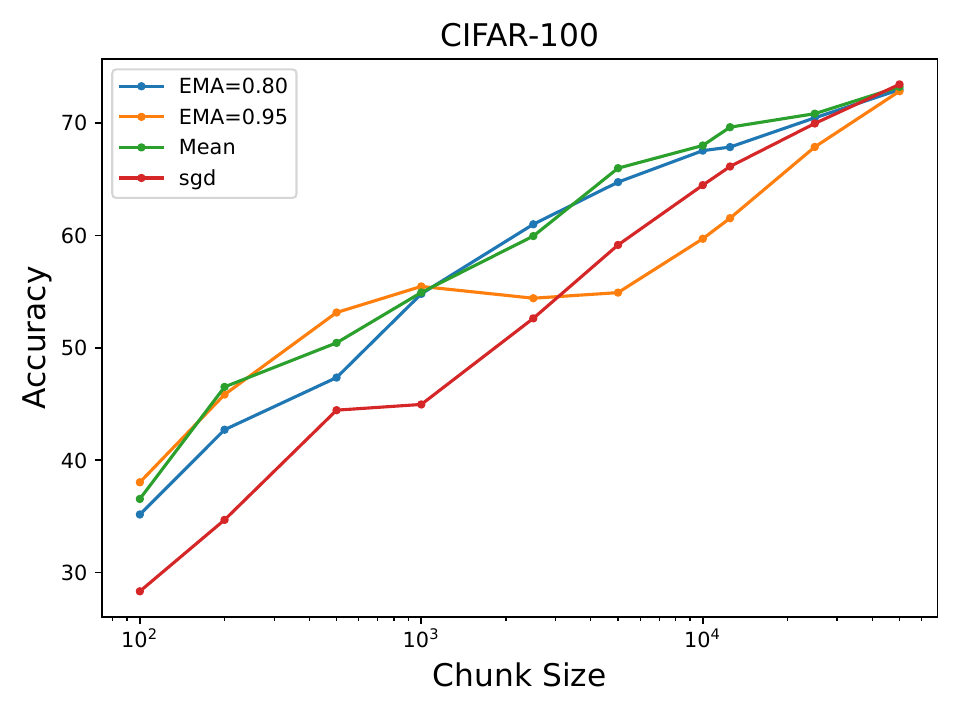} }}%
    \qquad
    \subfloat[]{{\includegraphics[trim={0 0 0 0mm}, clip, width=.5\linewidth]{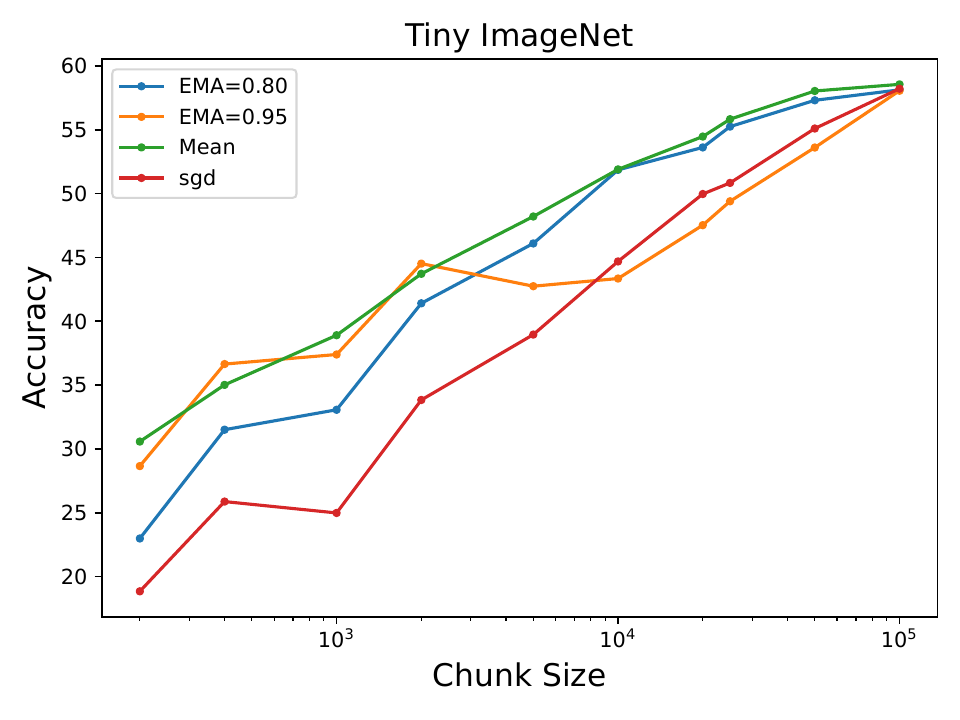} }}%
    \subfloat[]{{\includegraphics[trim={0 0 0 0mm}, clip, width=.5\linewidth]{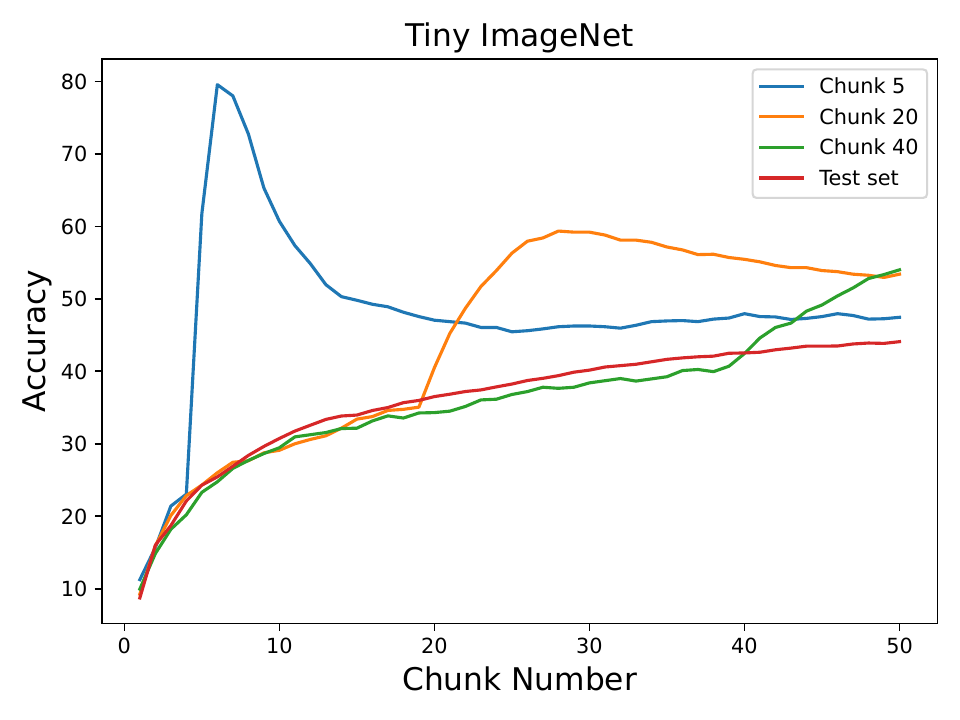} }}%
    \caption{Plots (a), (b) and (c) show the end-of-training accuracy when learning with the given chunk size for CIFAR-10, CIFAR-100 and Tiny ImageNet, where \emph{sgd} is learning without weight averaging and we display EMA results for $\alpha$$=$$0.8$ and $0.95$. These plots show that using weight averaging, in particular mean weight averaging, improves performance in the chunking setting. Plot (d) shows when using mean weight averaging the accuracy at the end of learning on each chunk for the training set of the $5^{th}$, $20^{th}$ and $40^{th}$ chunks and the test set, for Tiny ImageNet with 50 chunks, corresponding to a chunk size of 2000. The plot demonstrates, when compared to Figure~\ref{fig:50ChunkForgettingCurve}, that mean weight averaging forgets less than plain SGD training.}
    \label{fig:WeightAvgChunkingCurves}
\end{figure}
To explore if it is possible to improve performance in the chunking setting and whether this performance transfers to standard CL, we look at using per-chunk weight averaging. The reason we look at weight averaging is that by averaging over past weights the learner should better preserve information about previous chunks, reducing forgetting and hence improve performance. Also, in Appendix~\ref{appen:LinCase} we look at chunking when using linear models and show that weight averaging performs well. The simple weight averaging method we look at, \emph{per-chunk weight averaging}, consists of training the model as normal but we additionally store an average of the weights learnt at the end of each chunk. The per-chunk weight average is not used in training but in evaluation is used as the weights of the network. Here we consider the \emph{weights} to be all the parameters of the neural network, including batch normalisation statistics \citep{Ioffe2015Batch}. More specifically, we look at using in evaluation the mean or an exponential moving average (EMA) of the weights found after training on each chunk up to some chunk $k$, defined by
\begin{align}
    \thetaB^{MEAN}_{k} &= \frac{1}{k}\sum_{t=1}^{k}\thetaB_t \\
    \thetaB^{EMA}_{k}  &= \alpha\thetaB^{EMA}_{k-1}+(1-\alpha)\thetaB_k,
\end{align}
where $\thetaB_t$ is the value of the weights after learning on chunk $C_t$ and for EMA, $\alpha \in [0,1]$ controls how much weight is given to old versus newly learnt end-of-chunk weights. 

To observe whether per-chunk weight averaging improves performance in the chunking setting, we carry out experiments using it in combination with plain SGD training. The reason we only look at plain SGD training and not a CL method is that, as shown in Figures~\ref{fig:NNChunkingCurves} and \ref{fig:tinyimgChunkingCurve}, no CL method looked at performs any better than SGD in the chunking setting. The experimental setup is the same as the previous experiments and is described in Appendix~\ref{appen:expDetails}. The results of the experiments are presented in plots (a), (b) and (c) of Figure~\ref{fig:WeightAvgChunkingCurves} and show that it is clear that for all three datasets---CIFAR-10, CIFAR-100 and Tiny ImageNet---using a per-chunk weight average in evaluation increases accuracy. For instance, for the smallest chunk size looked at for each dataset, using mean weight averaging improves accuracy by $+4.32\%$, $+8.22\%$ and $+11.73\%$ for CIFAR-10, CIFAR-100 and Tiny ImageNet, respectively. Additionally, Figure~\ref{fig:WeightAvgChunkingCurves} demonstrates that using the mean is better than or comparable to using EMA for nearly all chunk sizes on each dataset. We only display EMA for two $\alpha$ values in the figure but we looked at many more in Appendix~\ref{appen:EMAChunkingCurves}, and selected the two best values to show in Figure~\ref{fig:WeightAvgChunkingCurves}. So, our results show that using the mean of the weights learnt after learning on each chunk for prediction is an effective way to improve performance in the chunking setting. 

To analyse why per-chunk weight averaging improves performance, we look at how well it preserves the information of past chunks. To do this, as in Figure~\ref{fig:50ChunkForgettingCurve}, we measure, for per-chunk mean weight averaging, the test accuracy and the accuracy on the training data of the $5^{th}$, $20^{th}$ and $40^{th}$ chunks at the end of learning on each chunk, when using 50 chunks. The results are shown in plot (d) of Figure~\ref{fig:WeightAvgChunkingCurves} for Tiny ImageNet and for CIFAR-10 and CIFAR-100 in Appendix~\ref{appen:add50ChunkForgettingCurve}. By comparing these results to the ones when using the final weights for evaluation, shown in Figure~\ref{fig:50ChunkForgettingCurve}, we see that when using per-chunk mean weight averaging more information is preserved from previous chunks. This is because using it gives higher accuracy on the training data from previous chunks than the test set long after that chunk was trained on. While, when using the final weights for evaluation this is not the case, as after learning on a chunk the accuracy on the training data of that chunk drops quickly down to around the test set accuracy. This suggests that part of the reason per-chunk weight averaging performs well is that it forgets less than plain SGD training in the chunking setting.      

\subsection{Application to Continual Learning}
\begingroup
\setlength{\tabcolsep}{2.5pt} 
\begin{table*}[t]
  \caption{Accuracy of CL methods in online and standard CL settings when using per-chunk weight averaging (WA-) or not, averaged over 3 runs and where we report the standard error over the runs. We also present results for IMM a CL weight averaging method, which performs worse than using per-chunk weight averaging with any base method. The table indicates that in general using weight averaging improves performance as shown by the positive performance improvement ($\Delta$Acc) when using weight averaging for almost all method/dataset/setting combinations.}
  \label{tab:CLWeightAvg}
  \centering
  \begin{tabular}{llllllll}
    \toprule
    & & \multicolumn{2}{c}{CIFAR-10} & \multicolumn{2}{c}{CIFAR-100} & \multicolumn{2}{c}{Tiny ImageNet} \\
    \cmidrule(r){3-4} \cmidrule(r){5-6} \cmidrule(r){7-8}
    Setting & Method & Class-IL & Task-IL  & Class-IL & Task-IL & Class-IL & Task-IL \\
    \midrule
    \multirow{ 14}{*}{Online} & DER++ & $34.76_{\pm 2.20}$ & $78.56_{\pm 1.10}$ & $6.73_{\pm 0.26}$ & $41.21_{\pm 1.34}$ & $5.48_{\pm 0.21}$ & $30.95_{\pm 0.11}$ \\
    & WA-DER++ & $33.46_{\pm 0.72}$ & $81.97_{\pm 0.25}$ & $12.34_{\pm 0.19}$ & $52.34_{\pm 0.43}$ & $8.53_{\pm 0.05}$ & $39.32_{\pm 0.55}$ \\
    & \Cg $\Delta$Acc ($\uparrow$)  & \Cg $-1.30$ & \Cg $+3.41$ & \Cg $+5.61$ & \Cg $+11.13$ & \Cg $+3.05$ & \Cg $+8.37$ \\
    \cmidrule(){2-8}
    & ER & $36.19_{\pm 1.19}$ & $81.89_{\pm 0.92}$  & $8.45_{\pm 0.45}$ & $44.14_{\pm 1.31}$ & $5.56_{\pm 0.21}$  & $27.23_{\pm 0.65}$  \\
    & WA-ER & $39.59_{\pm 0.60}$ & $84.27_{\pm 0.37}$ & $14.01_{\pm 0.23}$ & $50.66_{\pm 0.77}$ & $7.77_{\pm 0.09}$ & $34.26_{\pm 0.33}$ \\
    & \Cg $\Delta$Acc ($\uparrow$) & \Cg $+3.40$ & \Cg $+2.38$ & \Cg $+5.56$ & \Cg $+6.52$ & \Cg $+2.21$ & \Cg $+7.03$ \\
    \cmidrule(){2-8}
    & AGEM & $16.82_{\pm 0.61}$ & $70.70_{\pm 1.92}$ & $4.70_{\pm 0.51}$ & $29.56_{\pm 1.93}$ & $3.93_{\pm 0.22}$ & $20.53_{\pm 1.30}$ \\
    & WA-AGEM & $22.59_{\pm 1.04}$ & $72.37_{\pm 3.03}$ & $10.73_{\pm 0.35}$ & $44.68_{\pm 0.58}$ & $9.06_{\pm 0.47}$ & $34.44_{\pm 0.59}$ \\
    & \Cg$\Delta$Acc ($\uparrow$)  & \Cg $+5.77$ & \Cg $+1.67$ & \Cg $+6.03$ & \Cg $+20.39$ & \Cg $+5.13$ & \Cg $+13.91$ \\
    \cmidrule(){2-8}
    & GSS & $27.33_{\pm 1.26}$ & $81.28_{\pm 1.47}$ & $7.93_{\pm 0.16}$ & $49.95_{\pm 0.24}$ & $5.59_{\pm 0.11}$ & $36.00_{\pm 0.49}$ \\
    & WA-GSS & $35.03_{\pm 0.50}$ & $84.51_{\pm 0.41}$ & $8.40_{\pm 0.32}$ & $54.68_{\pm 0.28}$ & $4.82_{\pm 0.06}$ & $42.69_{\pm 0.38}$ \\
    & \Cg $\Delta$Acc ($\uparrow$) & \Cg $+7.70$ & \Cg $+3.23$ & \Cg $+0.47$ & \Cg $+4.73$ & \Cg $-0.77$ & \Cg $+6.69$ \\
    \cmidrule(){2-8}
    & IMM & $23.90_{\pm 1.40}$ & $68.84_{\pm 0.77}$ & $3.39_{\pm 0.29}$ & $20.65_{\pm 0.55}$ & $1.03_{\pm 0.08}$ & $9.69_{\pm 0.29}$ \\
    \midrule
    \multirow{ 14}{*}{Standard} & DER++ & $53.18_{\pm 0.87}$ & $88.90_{\pm 0.30}$ & $16.26_{\pm 1.22}$ & $58.92_{\pm 0.36}$ & $11.08_{\pm 0.38}$ & $34.26_{\pm 0.32}$\\
    & WA-DER++ & $49.88_{\pm 1.63}$ & $93.25_{\pm 0.33}$ & $23.46_{\pm 1.48}$ & $72.46_{\pm 1.08}$ & $12.39_{\pm 0.93}$ & $49.51_{\pm 0.69}$ \\
    & \Cg $\Delta$Acc ($\uparrow$) & \Cg $-3.30$ & \Cg $+4.35$ & \Cg $+7.20$ & \Cg $+13.54$ & \Cg $+1.31$ & \Cg $+15.25$ \\
    \cmidrule(){2-8}
    & ER & $40.01_{\pm0.81}$ & $89.79_{\pm 0.75}$  & $11.78_{\pm0.34}$ & $57.80_{\pm1.02}$ & $8.36_{\pm0.16}$  & $31.72_{\pm0.46}$  \\
    & WA-ER & $56.49_{\pm 0.87}$ & $94.28_{\pm 0.17}$ & $24.24_{\pm 0.64}$ & $70.07_{\pm 0.29}$ & $12.31_{\pm 0.19}$ & $46.71_{\pm 0.33}$  \\
    & \Cg $\Delta$Acc ($\uparrow$) & \Cg $+16.48$ & \Cg $+4.49$ & \Cg $+12.46$ & \Cg $+12.27$ & \Cg $+3.95$ & \Cg $+14.99$ \\
    \cmidrule(){2-8}
    & AGEM & $20.19_{\pm 0.28}$ & $85.80_{\pm 1.18}$ & $9.35_{\pm 0.01}$ & $46.99_{\pm 0.26}$ & $8.15_{\pm 0.05}$ & $24.76_{\pm 0.62}$ \\
    & WA-AGEM & $38.87_{\pm 2.83}$ & $92.06_{\pm 0.61}$ & $18.05_{\pm 0.68}$ & $65.23_{\pm 0.61}$ & $10.42_{\pm 0.32}$ & $42.75_{\pm 0.25}$ \\
    & \Cg $\Delta$Acc ($\uparrow$) & \Cg $+18.68$ & \Cg $+6.26$ & \Cg $+8.70$ & \Cg $+18.24$ & \Cg $+2.27$ & \Cg $+17.99$ \\
    \cmidrule(){2-8}
    & GSS &  $30.91_{\pm 1.02}$ & $86.08_{\pm 0.35}$ & $10.74_{\pm 0.10}$ & $50.30_{\pm 0.28}$ & $8.30_{\pm 0.01}$ & $27.55_{\pm 1.04}$ \\
    & WA-GSS & $51.58_{\pm 1.14}$ & $93.75_{\pm 0.43}$ & $14.78_{\pm 0.57}$ & $69.20_{\pm 0.35}$ & $6.13_{\pm 0.07}$ & $46.57_{\pm 1.16}$ \\
    & \Cg $\Delta$Acc ($\uparrow$) & \Cg $+20.67$ & \Cg $+7.67$ & \Cg $+4.04$ & \Cg $+18.90$ & \Cg $-2.17$ & \Cg $+19.02$ \\
    \cmidrule(){2-8}
    & IMM & $33.42_{\pm 1.97}$ & $89.91_{\pm 1.19}$ & $12.28_{\pm 1.33}$ & $43.46_{\pm 2.00}$ & $4.91_{\pm 0.58}$ & $22.28_{\pm 0.73}$ \\
    \bottomrule
  \end{tabular}
\end{table*}
\endgroup
While per-chunk weight averaging improves performance in the chunking setting, it is also important to see how this translates to the full CL setting, so that we can see how work on the chunking setting can impact CL in general. To do this we perform experiments using mean weight averaging in class and task incremental learning \citep{van2019three}, the two most common CL scenarios, using four standard well-performing methods: DER++ \citep{buzzega2020dark}, experience replay (ER) \citep{Chaudhry2020Continual}, AGEM \citep{Chaudhry2019Efficient} and GSS \citep{Aljundi2019Gradient}. The difference between class and task incremental learning is that at test time for task-incremental learning each method only predicts which class a data instance is between the classes of that data instance's task, while for class-incremental learning the method has to classify between all classes seen. As in common with the rest of this work and many works on continual learning \citep{Delange2021A, buzzega2020dark}, we use CIFAR-10, CIFAR-100 and Tiny ImageNet as the datasets for this experiment, splitting CIFAR-10 into 5 tasks each containing the data of 2 classes and splitting CIFAR-100 and Tiny ImageNet into 10 tasks each consisting of the data of 10 classes for CIFAR-100 and 20 for Tiny ImageNet. In addition to performing experiments using the standard CL setting (described in Section~\ref{sec:prelim}), we also present results for online CL \citep{mai2021online}. \emph{Online CL} is the same as standard CL but the leaner sees the data for each task as a sequence of mini-batches each of which is used to update the learner only once and is not revisited. For standard CL, methods can repeatedly iterate over the data of a task, in our experiments for each task we use 50 epochs for CIFAR-10 and CIFAR-100 and 100 epochs for Tiny ImageNet, like previous work \citep{buzzega2020dark} and we set the memory size to be 100 examples for all experiments.  

The results of the experiments on per-chunk mean weight averaging in CL are presented in Table~\ref{tab:CLWeightAvg} and demonstrate that in almost all cases it improves performance. For example, in the standard CL setting using per-chunk mean weight averaging improves performance on average by $+6.39\%$, $+11.11\%$, $+12.02\%$ and $+11.36\%$ for DER++, ER, AGEM and GSS, respectively. While in the online CL setting it improves performance on average by $+5.05\%$, $+4.52\%$, $+8.82\%$ and $+3.68\%$ for DER++, ER, AGEM and GSS, respectively. However, for class-incremental learning with DER++ on CIFAR-10 and GSS on Tiny ImageNet per-chunk mean weight averaging does worse than using the final learnt weights. But, as a method will have access to both options when using per-chunk mean weight averaging by validating the performance of each option it should be possible to pick the better one, avoiding any accuracy loss. For completeness, in Appendix~\ref{appen:EMACL} we also perform experiments with per-chunk EMA weight averaging in CL, showing that it performs worse than using the mean, like in the chunking setting. So, in summary, we have shown that per-chunk mean weight averaging improves performance in the chunking setting and that, in general, this improvement transfers to CL, showing that work on the chunking sub-problem can impact CL research as a whole.

We also perform experiments in standard and online CL with IMM \citep{lee2017overcoming} a popular weight averaging method for CL. IMM uses Fisher information to compute a weighted average of parameter weights. The results are presented in Table~\ref{tab:CLWeightAvg} and show that per-chunk weight averaging applied to any of the CL methods looked at outperforms IMM. So, while the main point of looking at per-chunk weight averaging is to demonstrate that performance can be improved in the chunking setting and that this performance transfers the standard CL setting; we also show that compared to IMM it is an effective weight averaging method for CL.

\section{Conclusions}
In this work we have looked at chunking, a sub-problem of continual learning (CL). We have presented results evidencing that it is responsible for a large part of the performance drop between offline and CL performance. Our results also reveal that current CL methods do not tackle the chunking problem, having comparable performance to plain SGD training in the chunking setting. Therefore, we have shown that chunking is a currently unaddressed problem which contributes a significant amount to the difficulty of CL. Additionally, we have demonstrated that a large amount of forgetting happens in the chunking setting, which casts into doubt the common belief that forgetting is caused mainly by task shift \citep{lee2021continual, ramasesh2020anatomy}. We also showed that performance on the chunking sub-problem can be improved---using per-chunk weight averaging. Furthermore, we demonstrated that this increased performance in the chunking setting transfers to the full CL setting, indicating that future work on chunking has the possibility of improving CL as a whole.


\subsubsection*{Acknowledgments}
This work was kindly supported by ARM and EPSRC through an iCASE PhD scholarship.

\FloatBarrier
\bibliography{References}

\begin{thebibliography}{59}
\providecommand{\natexlab}[1]{#1}
\providecommand{\url}[1]{\texttt{#1}}
\expandafter\ifx\csname urlstyle\endcsname\relax
  \providecommand{\doi}[1]{doi: #1}\else
  \providecommand{\doi}{doi: \begingroup \urlstyle{rm}\Url}\fi

\bibitem[Aljundi et~al.(2018)Aljundi, Kelchtermans, and Tuytelaars]{aljundi2018task}
Rahaf Aljundi, Klaas Kelchtermans, and Tinne Tuytelaars.
\newblock Task-{F}ree {C}ontinual {L}earning.
\newblock \emph{arXiv preprint arXiv:1812.03596}, 2018.

\bibitem[Aljundi et~al.(2019)Aljundi, Lin, Goujaud, and Bengio]{Aljundi2019Gradient}
Rahaf Aljundi, Min Lin, Baptiste Goujaud, and Yoshua Bengio.
\newblock Gradient {B}ased {S}ample {S}election for {O}nline {C}ontinual {L}earning.
\newblock In \emph{Proceedings of the 33rd Conference on the Advances in Neural Information Processing Systems}, pp.\  11816--11825, 2019.

\bibitem[Antoniou et~al.(2020)Antoniou, Patacchiola, Ochal, and Storkey]{antoniou2020defining}
Antreas Antoniou, Massimiliano Patacchiola, Mateusz Ochal, and Amos Storkey.
\newblock Defining {B}enchmarks for {C}ontinual {F}ew-shot {L}earning.
\newblock \emph{arXiv preprint arXiv:2004.11967}, 2020.

\bibitem[Ash \& Adams(2020)Ash and Adams]{ash2020warm}
Jordan Ash and Ryan~P Adams.
\newblock On {W}arm-{S}tarting {N}eural {N}etwork {T}raining.
\newblock In \emph{Proceedings of the 34th Conference on the Advances in Neural Information Processing Systems}, pp.\  3884--3894, 2020.

\bibitem[Bang et~al.(2021)Bang, Kim, Yoo, Ha, and Choi]{Bang2021Rainbow}
Jihwan Bang, Heesu Kim, YoungJoon Yoo, Jung-Woo Ha, and Jonghyun Choi.
\newblock Rainbow {M}emory: Continual {L}earning with a {M}emory of {D}iverse {S}amples.
\newblock In \emph{Proceedings of the 2021 IEEE/CVF Conference on Computer Vision and Pattern Recognition}, pp.\  8218--8227, 2021.

\bibitem[Bang et~al.(2022)Bang, Koh, Park, Song, Ha, and Choi]{bang2022online}
Jihwan Bang, Hyunseo Koh, Seulki Park, Hwanjun Song, Jung-Woo Ha, and Jonghyun Choi.
\newblock Online {C}ontinual {L}earning on a {C}ontaminated {D}ata {S}tream with {B}lurry {T}ask {B}oundaries.
\newblock In \emph{Proceedings of the 2022 IEEE/CVF Conference on Computer Vision and Pattern Recognition}, pp.\  9275--9284, 2022.

\bibitem[Boschini et~al.(2022)Boschini, Bonicelli, Buzzega, Porrello, and Calderara]{boschini2022class}
Matteo Boschini, Lorenzo Bonicelli, Pietro Buzzega, Angelo Porrello, and Simone Calderara.
\newblock Class-{I}ncremental {C}ontinual {L}earning into the {E}xtended {DER}-verse.
\newblock \emph{IEEE Transactions on Pattern Analysis and Machine Intelligence}, 45\penalty0 (5):\penalty0 5497--5512, 2022.

\bibitem[Bottou \& LeCun(2003)Bottou and LeCun]{bottou2003large}
L{\'e}on Bottou and Yann LeCun.
\newblock Large {S}cale {O}nline {L}earning.
\newblock In \emph{Proceedings of the 17th Conference on the Advances in Neural Information Processing Systems}, 2003.

\bibitem[Buzzega et~al.(2020)Buzzega, Boschini, Porrello, Abati, and Calderara]{buzzega2020dark}
Pietro Buzzega, Matteo Boschini, Angelo Porrello, Davide Abati, and Simone Calderara.
\newblock Dark {E}xperience for {G}eneral {C}ontinual {L}earning: A {S}trong, {S}imple {B}aseline.
\newblock In \emph{Proceedings of the 34th Conference on the Advances in Neural Information Processing Systems}, pp.\  15920--15930, 2020.

\bibitem[Caccia et~al.(2021)Caccia, Aljundi, Asadi, Tuytelaars, Pineau, and Belilovsky]{caccia2021new}
Lucas Caccia, Rahaf Aljundi, Nader Asadi, Tinne Tuytelaars, Joelle Pineau, and Eugene Belilovsky.
\newblock New {I}nsights on {R}educing {A}brupt {R}epresentation {C}hange in {O}nline {C}ontinual {L}earning.
\newblock In \emph{Proceedings of the 10th International Conference on Learning Representations}, 2021.

\bibitem[Caccia et~al.(2022)Caccia, Xu, Ott, Ranzato, and Denoyer]{caccia2022anytime}
Lucas Caccia, Jing Xu, Myle Ott, Marcaurelio Ranzato, and Ludovic Denoyer.
\newblock On {A}nytime {L}earning at {M}acroscale.
\newblock In \emph{Proceedings of the 1st Conference on Lifelong Learning Agents}, pp.\  165--182, 2022.

\bibitem[Caccia et~al.(2020)Caccia, Rodriguez, Ostapenko, Normandin, Lin, Page-Caccia, Laradji, Rish, Lacoste, and V{\'a}zquez]{caccia2020online}
Massimo Caccia, Pau Rodriguez, Oleksiy Ostapenko, Fabrice Normandin, Min Lin, Lucas Page-Caccia, Issam~Hadj Laradji, Irina Rish, Alexandre Lacoste, and David V{\'a}zquez.
\newblock Online {F}ast {A}daptation and {K}nowledge {A}ccumulation ({OSAKA}): A new {A}pproach to {C}ontinual {L}earning.
\newblock In \emph{Proceedings of the 34th Conference on the Advances in Neural Information Processing Systems}, pp.\  16532--16545, 2020.

\bibitem[Chaudhry et~al.(2019)Chaudhry, Ranzato, Rohrbach, and Elhoseiny]{Chaudhry2019Efficient}
Arslan Chaudhry, Marc’Aurelio Ranzato, Marcus Rohrbach, and Mohamed Elhoseiny.
\newblock Efficient {L}ifelong {L}earning with {A-GEM}.
\newblock In \emph{Proceedings of the 7th International Conference on Learning Representations}, 2019.

\bibitem[Chaudhry et~al.(2020)Chaudhry, Khan, Dokania, and Torr]{Chaudhry2020Continual}
Arslan Chaudhry, Naeemullah Khan, Puneet Dokania, and Philip Torr.
\newblock Continual {L}earning in {L}ow-rank {O}rthogonal {S}ubspaces.
\newblock In \emph{Proceedings of the 34th Conference on the Advances in Neural Information Processing Systems}, pp.\  9900--9911, 2020.

\bibitem[de~Angulo \& Torras(1995)de~Angulo and Torras]{deAgulo95}
V.R. de~Angulo and C.~Torras.
\newblock On-{L}ine {L}earning with {M}inimal {D}egradation in {F}eedforward {N}etworks.
\newblock \emph{IEEE Transactions on Neural Networks}, 6\penalty0 (3):\penalty0 657--668, 1995.

\bibitem[De~Lange et~al.(2023)De~Lange, van~de Ven, and Tuytelaars]{de2023continual}
Matthias De~Lange, Gido van~de Ven, and Tinne Tuytelaars.
\newblock Continual {E}valuation for {L}ifelong {L}earning: Identifying the {S}tability {G}ap.
\newblock In \emph{Proceedings of the 11th International Conference on Learning Representations}, 2023.

\bibitem[Delange et~al.(2021)Delange, Aljundi, Masana, Parisot, Jia, Leonardis, Slabaugh, and Tuytelaars]{Delange2021A}
Matthias Delange, Rahaf Aljundi, Marc Masana, Sarah Parisot, Xu~Jia, Ales Leonardis, Greg Slabaugh, and Tinne Tuytelaars.
\newblock A {C}ontinual {L}earning {S}urvey: {D}efying {F}orgetting in {C}lassification {T}asks.
\newblock \emph{IEEE Transactions on Pattern Analysis and Machine Intelligence}, 2021.

\bibitem[Garg et~al.(2023)Garg, Farajtabar, Pouransari, Vemulapalli, Mehta, Tuzel, Shankar, and Faghri]{garg2023tic}
Saurabh Garg, Mehrdad Farajtabar, Hadi Pouransari, Raviteja Vemulapalli, Sachin Mehta, Oncel Tuzel, Vaishaal Shankar, and Fartash Faghri.
\newblock T{IC}-{CLIP}: Continual {T}raining of {CLIP} {M}odels.
\newblock \emph{arXiv preprint arXiv:2310.16226}, 2023.

\bibitem[Grossberg(1988)]{grossberg88}
Stephen Grossberg.
\newblock Nonlinear {N}eural {N}etworks: {P}rinciples, {M}echanisms and {A}rchitectures.
\newblock \emph{Neural Networks}, 1\penalty0 (1):\penalty0 17--61, 1988.

\bibitem[Hayes \& Kanan(2020)Hayes and Kanan]{hayes2020lifelong}
Tyler~L Hayes and Christopher Kanan.
\newblock Lifelong {M}achine {L}earning with {D}eep {S}treaming {L}inear {D}iscriminant {A}nalysis.
\newblock In \emph{Proceedings of the 2020 IEEE/CVF Conference on Computer Vision and Pattern Recognition Workshops}, pp.\  220--221, 2020.

\bibitem[He et~al.(2016)He, Zhang, Ren, and Sun]{he2016deep}
Kaiming He, Xiangyu Zhang, Shaoqing Ren, and Jian Sun.
\newblock Deep {R}esidual {L}earning for {I}mage {R}ecognition.
\newblock In \emph{Proceedings of the 2016 IEEE Conference on Computer Vision and Pattern Recognition}, pp.\  770--778, 2016.

\bibitem[Hoi et~al.(2021)Hoi, Sahoo, Lu, and Zhao]{Hoi202Online}
Steven~C.H. Hoi, Doyen Sahoo, Jing Lu, and Peilin Zhao.
\newblock Online {L}earning: {A} {C}omprehensive {S}urvey.
\newblock \emph{Neurocomputing}, 459:\penalty0 249--289, 2021.

\bibitem[Hsu et~al.(2018)Hsu, Liu, Ramasamy, and Kira]{hsu2018re}
Yen-Chang Hsu, Yen-Cheng Liu, Anita Ramasamy, and Zsolt Kira.
\newblock Re-evaluating {C}ontinual {L}earning {S}cenarios: A {C}ategorization and {C}ase for {S}trong {B}aselines.
\newblock In \emph{Proceedings of the 3rd Continual Learning Workshop, at the 32nd Conference on the Advances in Neural Information Processing Systems}, 2018.

\bibitem[Ioffe \& Szegedy(2015)Ioffe and Szegedy]{Ioffe2015Batch}
Sergey Ioffe and Christian Szegedy.
\newblock Batch {N}ormalization: Accelerating {D}eep {N}etwork {T}raining by {R}educing {I}nternal {C}ovariate {S}hift.
\newblock In \emph{Proceedings of the 32nd International Conference on Machine Learning}, pp.\  448--456, 2015.

\bibitem[Izmailov et~al.(2018)Izmailov, Podoprikhin, Garipov, Vetrov, and Wilson]{izmailov2018averaging}
Pavel Izmailov, Dmitrii Podoprikhin, Timur Garipov, Dmitry Vetrov, and Andrew~Gordon Wilson.
\newblock Averaging {W}eights leads to {W}ider {O}ptima and better {G}eneralization.
\newblock \emph{arXiv preprint arXiv:1803.05407}, 2018.

\bibitem[Kirkpatrick et~al.(2017)Kirkpatrick, Pascanu, Rabinowitz, Veness, Desjardins, Rusu, Milan, Quan, Ramalho, and Grabska-Barwinska]{kirkpatrick2017overcoming}
James Kirkpatrick, Razvan Pascanu, Neil Rabinowitz, Joel Veness, Guillaume Desjardins, Andrei~A Rusu, Kieran Milan, John Quan, Tiago Ramalho, and Agnieszka Grabska-Barwinska.
\newblock Overcoming {C}atastrophic {F}orgetting in {N}eural {N}etworks.
\newblock \emph{Proceedings of the National Academy of Sciences}, 114\penalty0 (13):\penalty0 3521--3526, 2017.

\bibitem[Krizhevsky(2009)]{krizhevsky2009learning}
Alex Krizhevsky.
\newblock Learning {M}ultiple {L}ayers of {F}eatures from {T}iny {I}mages.
\newblock \emph{Preprint}, 2009.

\bibitem[Lee et~al.(2020)Lee, Joo, Hong, and Kim]{lee2020residual}
Janghyeon Lee, Donggyu Joo, Hyeong~Gwon Hong, and Junmo Kim.
\newblock Residual {C}ontinual {L}earning.
\newblock In \emph{Proceedings of the 34th AAAI Conference on Artificial Intelligence}, pp.\  4553--4560, 2020.

\bibitem[Lee et~al.(2017)Lee, Kim, Jun, Ha, and Zhang]{lee2017overcoming}
Sang-Woo Lee, Jin-Hwa Kim, Jaehyun Jun, Jung-Woo Ha, and Byoung-Tak Zhang.
\newblock Overcoming {C}atastrophic {F}orgetting by {I}ncremental {M}oment {M}atching.
\newblock In \emph{Proceedings of the 31st Conference on the Advances in Neural Information Processing Systems}, 2017.

\bibitem[Lee et~al.(2021)Lee, Goldt, and Saxe]{lee2021continual}
Sebastian Lee, Sebastian Goldt, and Andrew Saxe.
\newblock Continual {L}earning in the {T}eacher-{S}tudent {S}etup: {I}mpact of {T}ask {S}imilarity.
\newblock In \emph{Proceedings of the 38th International Conference on Machine Learning}, pp.\  6109--6119, 2021.

\bibitem[Lee \& Storkey(2024)Lee and Storkey]{lee2023class}
Thomas~L Lee and Amos Storkey.
\newblock Approximate {B}ayesian {C}lass-{C}onditional {M}odels under {C}ontinuous {R}epresentation {S}hift.
\newblock In \emph{Proceedings of the 27th International Conference on Artificial Intelligence and Statistics}, pp.\  3628--3636, 2024.

\bibitem[Lesort et~al.(2022)Lesort, Ostapenko, Misra, Arefin, Rodr{\'\i}guez, Charlin, and Rish]{lesort2023challenging}
Timoth{\'e}e Lesort, Oleksiy Ostapenko, Diganta Misra, Md~Rifat Arefin, Pau Rodr{\'\i}guez, Laurent Charlin, and Irina Rish.
\newblock Challenging common assumptions about catastrophic forgetting.
\newblock \emph{arXiv preprint arXiv:2207.04543}, 2022.

\bibitem[Lin et~al.(2022)Lin, Yang, Fan, and Zhang]{lin2022beyond}
Sen Lin, Li~Yang, Deliang Fan, and Junshan Zhang.
\newblock Beyond not-{F}orgetting: {C}ontinual {L}earning with {B}ackward {K}nowledge {T}ransfer.
\newblock In \emph{Procceding of the 36th Conference on the Advances in Neural Information Processing Systems}, pp.\  16165--16177, 2022.

\bibitem[Lomonaco \& Maltoni(2017)Lomonaco and Maltoni]{lomonaco2017core50}
Vincenzo Lomonaco and Davide Maltoni.
\newblock Core50: A {N}ew {D}ataset and {B}enchmark for {C}ontinuous {O}bject {R}ecognition.
\newblock In \emph{Proceedings of the 1st Conference on Robot Learning}, pp.\  17--26, 2017.

\bibitem[Mai et~al.(2021)Mai, Li, Jeong, Quispe, Kim, and Sanner]{mai2021online}
Zheda Mai, Ruiwen Li, Jihwan Jeong, David Quispe, Hyunwoo Kim, and Scott Sanner.
\newblock Online {C}ontinual {L}earning in {I}mage {C}lassification: An {E}mpirical {S}urvey.
\newblock \emph{arXiv preprint arXiv:2101.10423}, 2021.

\bibitem[Maji et~al.(2013)Maji, Rahtu, Kannala, Blaschko, and Vedaldi]{maji2013fine}
Subhransu Maji, Esa Rahtu, Juho Kannala, Matthew Blaschko, and Andrea Vedaldi.
\newblock Fine-grained {V}isual {C}lassification of {A}ircraft.
\newblock \emph{arXiv preprint arXiv:1306.5151}, 2013.

\bibitem[Mi et~al.(2020)Mi, Kong, Lin, Yu, and Faltings]{mi2020generalized}
Fei Mi, Lingjing Kong, Tao Lin, Kaicheng Yu, and Boi Faltings.
\newblock Generalized {C}lass {I}ncremental {L}earning.
\newblock In \emph{Proceedings of the 2020 IEEE/CVF Conference on Computer Vision and Pattern Recognition Workshops}, pp.\  240--241, 2020.

\bibitem[Minka(2000)]{minka2000bayesian}
Thomas Minka.
\newblock Bayesian {L}inear {R}egression.
\newblock Technical report, Microsoft Research, 2000.

\bibitem[Mirzadeh et~al.(2020)Mirzadeh, Farajtabar, Pascanu, and Ghasemzadeh]{Mirzadeh2020Understanding}
Seyed~Iman Mirzadeh, Mehrdad Farajtabar, Razvan Pascanu, and Hassan Ghasemzadeh.
\newblock Understanding the {R}ole of {T}raining {R}egimes in {C}ontinual {L}earning.
\newblock In \emph{Proceedings of the 34th Conference on the Advances in Neural Information Processing Systems}, pp.\  7308--7320, 2020.

\bibitem[Nakkiran et~al.(2021)Nakkiran, Kaplun, Bansal, Yang, Barak, and Sutskever]{nakkiran2021deep}
Preetum Nakkiran, Gal Kaplun, Yamini Bansal, Tristan Yang, Boaz Barak, and Ilya Sutskever.
\newblock Deep {D}ouble {D}escent: Where {B}igger {M}odels and {M}ore {D}ata {H}urt.
\newblock \emph{Journal of Statistical Mechanics: Theory and Experiment}, 2021\penalty0 (12):\penalty0 124003, 2021.

\bibitem[Ostapenko et~al.(2022)Ostapenko, Lesort, Rodr{\'\i}guez, Arefin, Douillard, Rish, and Charlin]{ostapenko2022foundational}
Oleksiy Ostapenko, Timothee Lesort, Pau Rodr{\'\i}guez, Md~Rifat Arefin, Arthur Douillard, Irina Rish, and Laurent Charlin.
\newblock Foundational {M}odels for {C}ontinual {L}earning: {A}n {E}mpirical {S}tudy of {L}atent {R}eplay.
\newblock \emph{arXiv preprint arXiv:2205.00329}, 2022.

\bibitem[Parisi et~al.(2019)Parisi, Kemker, Part, Kanan, and Wermter]{Parisi2019review}
German~I. Parisi, Ronald Kemker, Jose~L. Part, Christopher Kanan, and Stefan Wermter.
\newblock Continual {L}ifelong {L}earning with {N}eural {N}etworks: A review.
\newblock \emph{Neural Networks}, 113:\penalty0 54 -- 71, 2019.

\bibitem[Pelosin(2022)]{pelosin2022simpler}
Francesco Pelosin.
\newblock Simpler is {B}etter: off-the-shelf {C}ontinual {L}earning {T}hrough {P}retrained {B}ackbones.
\newblock \emph{arXiv preprint arXiv:2205.01586}, 2022.

\bibitem[Polikar et~al.(2001)Polikar, Upda, Upda, and Honavar]{polikar01}
R.~Polikar, L.~Upda, S.S. Upda, and V.~Honavar.
\newblock Learn++: An {I}ncremental {L}earning {A}lgorithm for {S}upervised {N}eural {N}etworks.
\newblock \emph{IEEE Transactions on Systems, Man, and Cybernetics, Part C (Applications and Reviews)}, 31\penalty0 (4):\penalty0 497--508, 2001.

\bibitem[Prabhu et~al.(2020)Prabhu, Torr, and Dokania]{prabhu2020gdumb}
Ameya Prabhu, Philip~HS Torr, and Puneet~K Dokania.
\newblock G{D}umb: A {S}imple {A}pproach that {Q}uestions our {P}rogress in {C}ontinual {L}earning.
\newblock In \emph{Procceding of the 16th European Conference on Computer Vision}, pp.\  524--540, 2020.

\bibitem[Prabhu et~al.(2023)Prabhu, Al~Kader~Hammoud, Dokania, Torr, Lim, Ghanem, and Bibi]{prabhu2023computationally}
Ameya Prabhu, Hasan~Abed Al~Kader~Hammoud, Puneet~K Dokania, Philip~HS Torr, Ser-Nam Lim, Bernard Ghanem, and Adel Bibi.
\newblock Computationally {B}udgeted {C}ontinual {L}earning: What {D}oes {M}atter?
\newblock In \emph{Proceedings of the 2023 IEEE/CVF Conference on Computer Vision and Pattern Recognition}, pp.\  3698--3707, 2023.

\bibitem[Ramasesh et~al.(2020)Ramasesh, Dyer, and Raghu]{ramasesh2020anatomy}
Vinay~V Ramasesh, Ethan Dyer, and Maithra Raghu.
\newblock Anatomy of {C}atastrophic {F}orgetting: {H}idden {R}epresentations and {T}ask {S}emantics.
\newblock \emph{arXiv preprint arXiv:2007.07400}, 2020.

\bibitem[Sahoo et~al.(2017)Sahoo, Pham, Lu, and Hoi]{sahoo2017online}
Doyen Sahoo, Quang Pham, Jing Lu, and Steven~CH Hoi.
\newblock Online {D}eep {L}earning: {L}earning {D}eep {N}eural {N}etworks on the {F}ly.
\newblock \emph{arXiv preprint arXiv:1711.03705}, 2017.

\bibitem[Stojanovski et~al.(2022)Stojanovski, Roth, and Akata]{stojanovski2022momentum}
Zafir Stojanovski, Karsten Roth, and Zeynep Akata.
\newblock Momentum-{B}ased {W}eight {I}nterpolation of {S}trong {Z}ero-{S}hot {M}odels for {C}ontinual {L}earning.
\newblock \emph{arXiv preprint arXiv:2211.03186}, 2022.

\bibitem[Storkey(1997)]{storkey1997increasing}
Amos Storkey.
\newblock Increasing the {C}apacity of a {H}opfield {N}etwork {W}ithout {S}acrificing {F}unctionality.
\newblock In \emph{Proceedings of the 7th International Conference on Artificial Neural Networks}, pp.\  451--456, 1997.

\bibitem[Tarvainen \& Valpola(2017)Tarvainen and Valpola]{tarvainen2017mean}
Antti Tarvainen and Harri Valpola.
\newblock Mean {T}eachers are {B}etter {R}ole {M}odels: {W}eight-{A}veraged {C}onsistency {T}argets {I}mprove {S}emi-{S}upervised {D}eep {L}earning {R}esults.
\newblock \emph{Proceedings of the 31st Conference on the Advances in Neural Information Processing Systems}, 2017.

\bibitem[Tishby \& Zaslavsky(2015)Tishby and Zaslavsky]{tishby2015deep}
Naftali Tishby and Noga Zaslavsky.
\newblock Deep {L}earning and the {I}nformation {B}ottleneck {P}rinciple.
\newblock In \emph{Proceedings of the 2015 IEEE information theory workshop}, pp.\  1--5. IEEE, 2015.

\bibitem[van~de Ven \& Tolias(2019)van~de Ven and Tolias]{van2019three}
Gido~M van~de Ven and Andreas~S Tolias.
\newblock Three {S}cenarios for {C}ontinual {L}earning.
\newblock \emph{arXiv preprint arXiv:1904.07734}, 2019.

\bibitem[Wainwright(2019)]{wainwright2019high}
Martin~J Wainwright.
\newblock \emph{High-{D}imensional {S}tatistics: A {N}on-{A}symptotic {V}iewpoint}.
\newblock 2019.

\bibitem[Wang et~al.(2023)Wang, Zhang, Su, and Zhu]{wang2023comprehensive}
Liyuan Wang, Xingxing Zhang, Hang Su, and Jun Zhu.
\newblock A {C}omprehensive {S}urvey of {C}ontinual {L}earning: {T}heory, {M}ethod and {A}pplication.
\newblock \emph{arXiv preprint arXiv:2302.00487}, 2023.

\bibitem[Wightman et~al.(2021)Wightman, Touvron, and Jegou]{wightman2021resnet}
Ross Wightman, Hugo Touvron, and Herve Jegou.
\newblock Res{N}et {S}trikes {B}ack: {A}n {I}mproved {T}raining {P}rocedure in timm.
\newblock In \emph{Proccedings of the Workshop on ImageNet: Past, Present, and Future, at the 35th Conference on the Advances in Neural Information Processing Systems}, 2021.

\bibitem[Wu et~al.(2015)Wu, Zhang, and Xu]{Stanford2015Tiny}
Jiayu Wu, Qixiang Zhang, and Guoxi Xu.
\newblock Tiny {I}magenet {C}hallenge (cs231n), http://tiny-imagenet.herokuapp.com/.
\newblock Technical report, Stanford, 2015.

\bibitem[Wu et~al.(2022)Wu, Caccia, Li, Li, Qi, and Haffari]{wu2022pretrained}
Tongtong Wu, Massimo Caccia, Zhuang Li, Yuan-Fang Li, Guilin Qi, and Gholamreza Haffari.
\newblock Pretrained {L}anguage {M}odels in {C}ontinual {L}earning: A {C}omparative {S}tudy.
\newblock In \emph{Proceedings of the 10th International Conference on Learning Representations}, 2022.

\bibitem[Zhang et~al.(2021)Zhang, Xie, Bai, Yu, Li, and Gao]{zhang2021survey}
Chen Zhang, Yu~Xie, Hang Bai, Bin Yu, Weihong Li, and Yuan Gao.
\newblock A {S}urvey on {F}ederated {L}earning.
\newblock \emph{Knowledge-Based Systems}, 216:\penalty0 106775, 2021.

\end{thebibliography}
\bibliographystyle{collas2024_conference}

\newpage
\appendix

\section{Limitations}
While we have aimed to be thorough in our analysis of the chunking problem, like any work there are limitations to what is presented in the paper. We list a number of limitations here. First, this work is mainly empirically based so it would be interesting to see what theoretical approaches can tell us about the chunking sub-problem and its relation CL. For example, the performance in the chunking setting is an approximate upper-bound to CL performance, as it a reduced and simpler setting due to removing the task-shift element. However, in this work we did not make the nature of the upper-bounding exact and so it would be useful to see if a theorem could be derived to formally express how chunking performance upper-bounds CL performance. Additionally, it would be interesting to see a theoretical approach to decomposing the difficulty of CL into its chunking and task-shift components. Another limitation is that while mean weight averaging does improve performance in the chunking setting, there is still a large gap between it and offline learning performance, for a small enough chunk size. This means there is space for better methods to be developed which better solve the chunking problem and hence hopefully also the full CL setting with task shift. Last, we only look at supervised CL in this work however the chunking problem is also a sub-problem of any self or semi-supervised CL setting. Therefore it would be interesting to see how much chunking contributes the difficulty of those settings and whether CL methods developed for those settings currently tackle the chunking problem.        

\section{Experimental Details}
\label{appen:expDetails}
For all of our results we follow the experimental protocol of \citet{buzzega2020dark} and \citet{boschini2022class}, and use a modification of the CL library \emph{Mammoth} used in those works to run the experiments. Therefore, for all our experiments we use a ResNet18 \citep{he2016deep} as the backbone model. Additionally, we utilize augmentations, applying random crops and horizontal flips to images trained on for all the datasets used. To be able to have a fair comparison in our chunking experiments all methods are trained using SGD and with the same number of epochs: 50 epochs for each chunk for CIFAR-10 and CIFAR-100 and 100 for Tiny ImageNet. We use the same mini-batch size for all experiments, which is 32 examples, and for replay CL methods we use 32 as the replay batch size as well. The hyperparameters of methods were found using a grid search on a validation set and are the same as in \citet{buzzega2020dark} and \citet{boschini2022class}; however, all the results were newly computed by the authors for this work. For all results on the chunking setting a learning rate of 0.1 was used to ensure a fair comparison between methods and chunk sizes. Last, the full list of CL methods evaluated in the chunking setting is: AGEM \citep{Chaudhry2019Efficient}, DER++ \citep{buzzega2020dark}, ER \citep{Chaudhry2020Continual}, ER-ACE \citep{caccia2021new}, EWC \citep{kirkpatrick2017overcoming}, GSS \citep{Aljundi2019Gradient} and plain SGD training. Plain SGD training is when the neural network is trained with SGD using the standard cross entropy loss, as in previous work \citep{buzzega2020dark, boschini2022class}.

\section{Positive Transfer and the Chunking Problem}
\label{appen:posTransfer}
While a large part of the effort in CL has thus far been to reduce forgetting, another key problem is improve the positive transfer capabilities of methods \citep{lin2022beyond}. \emph{Positive transfer} is the ability of a learner to use the data in the currently accessible chunk/task to to improve its knowledge of previous chunks/tasks and future chunks/tasks to be seen. It has been shown that current CL methods do not perform positive transfer well and that improving positive transfer capabilities will improve the performance of CL algorithms in general \citep{lee2023class, lin2022beyond}. While positive transfer has previously been explored between chunks/tasks of differing data distributions, our results on the chunking setting show that current CL methods fail to perform positive transfer well even if the chunks are all sampled from the same distribution. This is because, Figures~\ref{fig:NNChunkingCurves} and \ref{fig:tinyimgChunkingCurve} show that CL methods perform comparable to plain SGD training in the chunking setting and that Figure~\ref{fig:50ChunkForgettingCurve} shows that plain SGD training is very bad at positive transfer, as it looses instead of gains performance on past chunks when training on additional chunks. Therefore, we propose that improving performance in the chunking setting should lead to methods that are better at positive transfer and vice versa. Additionally, as the chunking setting is much simpler than other CL settings, being a sub-problem of all of them, hopefully by looking at this setting its simplicity will be a benefit when working on improving the positive transfer capabilities of methods. 

\section{The Chunking Problem in Online CL}
\label{appen:onlineCL}
In Section~\ref{sec:Analysis} we show that the chunking problem is a significant challenge in standard CL and here we describe why it is also a challenge for online CL. Unlike this work and past work in standard CL, most work in online CL compare not to offline performance but to an i.i.d upperbound. The \emph{i.i.d upperbound} is plain SGD training run in the online CL setting but all the mini-batches are drawn i.i.d without replacement from the dataset. Therefore, this upperbound computes the chunking performance of plain SGD for a chunk size of one mini-batch and where a method only takes a single update on each chunk (to the best of our knowledge this has not been explicitly discussed in the online CL literature and the chunking problem has remained unexamined until this work). Interestingly, online CL methods are now getting comparable performance to this upperbound \citep{caccia2021new, lee2023class}. This suggests that most of the remaining performance to be gained in online CL is in solving the chunking problem. Therefore, there is a large space in the literature to solve the chunking problem and in doing so improve online CL performance beyond this current i.i.d upperbound performance. In this paper we give the first steps towards this by showing we can improve on the i.i.d uppperbound performance by using per-chunk weight averaging (see Figure~\ref{fig:WeightAvgChunkingCurves} and Table~\ref{tab:CLWeightAvg}).

\section{The Stability Gap in the Chunking Setting}
\label{Appen:StabilityGap}
\begin{figure}[h]
    \centering
    \subfloat{{\includegraphics[width=.5\linewidth]{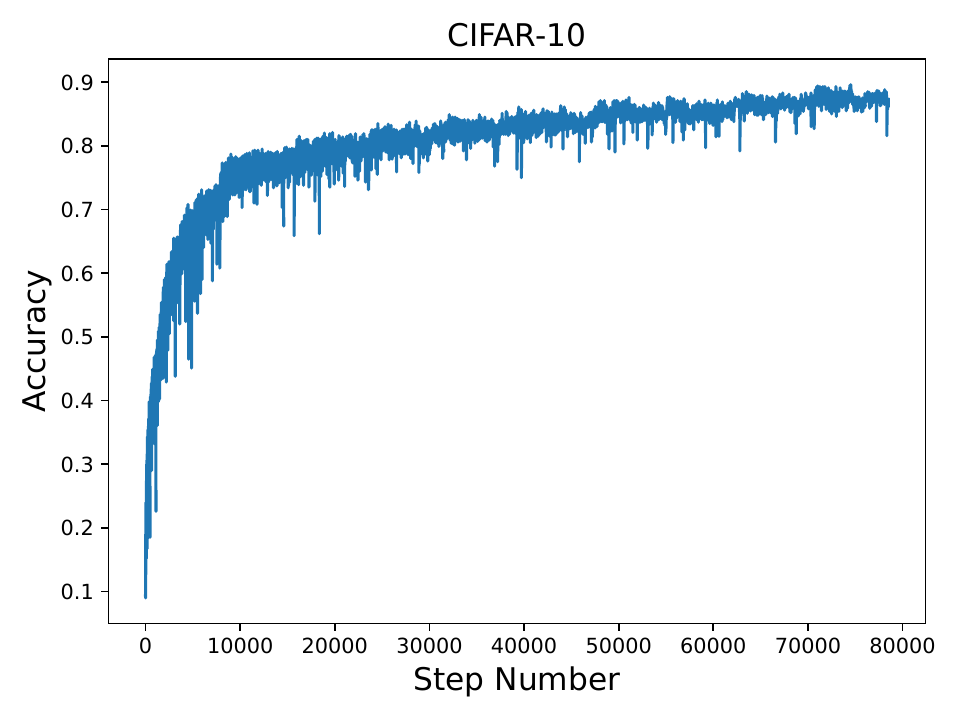} }}%
    \subfloat{{\includegraphics[width=.5\linewidth]{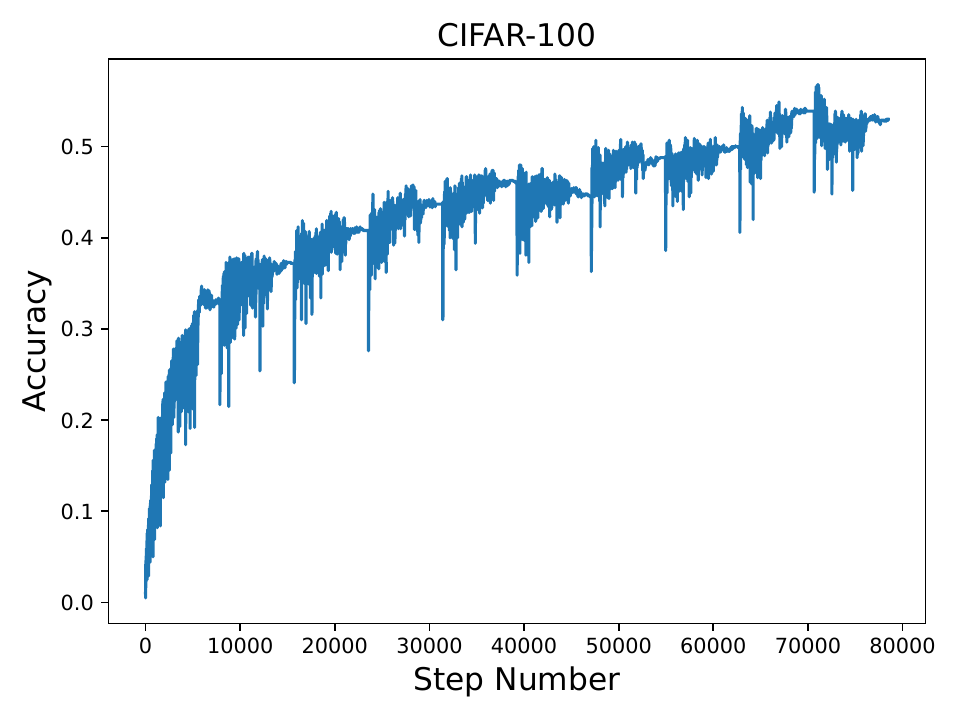} }}%
    \qquad
    \subfloat{{\includegraphics[width=.5\linewidth]{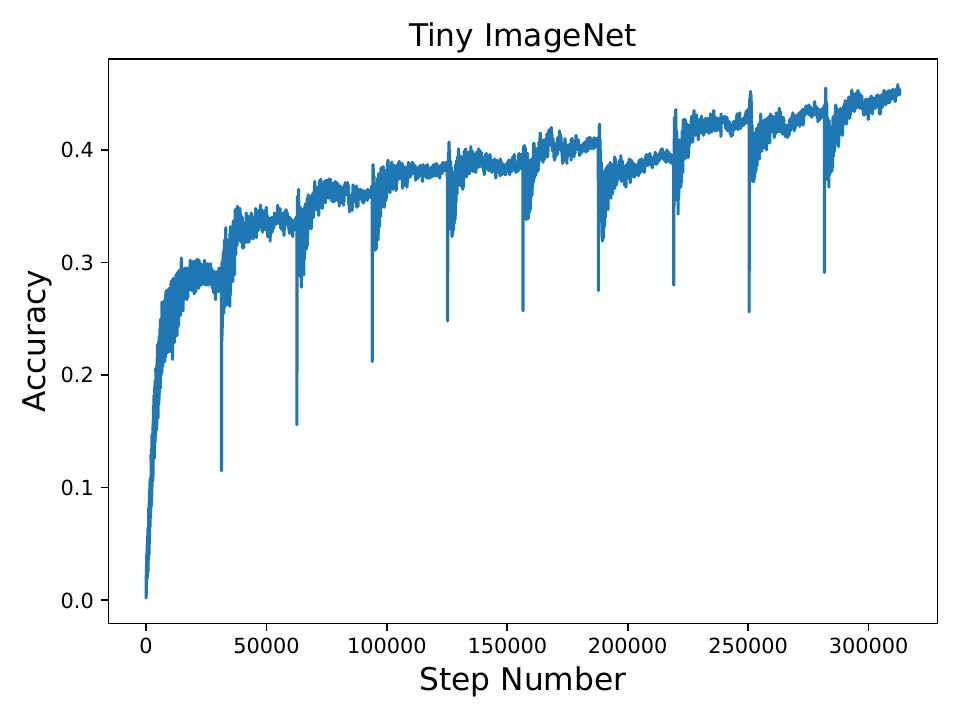} }}%
    \caption{Test accuracy curves for plain SGD training using 10 chunks on CIFAR-10, CIFAR-100 and Tiny ImageNet. Step number refers to the number of parameter updates which have been performed up to that point. The plots, apart from CIFAR-10, show that at the start of each chunk there is a large drop in test accuracy.}%
    \label{fig:SGDStabilityGap}%
\end{figure}
The stability gap is a recently discovered phenomena in standard CL \citep{de2023continual}, we show in this appendix that it also occurs in the chunking setting. The \emph{stability gap} is the phenomena that at the start of learning a new task the accuracy for previous tasks drops quickly by a significant amount and then recovers back to a stable level. This stable level is somewhat lower than the accuracy value before seeing the new task. The reason this is thought to happen \citep{de2023continual, caccia2021new} is that at the start of learning on a new task the performance on it will be poor, leading to large gradients and changes to the weights of the network. This potentially induces a drop in performance on previous tasks as the large weight updates remove information about previous tasks. Finally, as the weights change the performance on data stored in memory from previous tasks will reduce, increasingly impacting the parameter updates to perform better on previous tasks, recovering a large amount of the performance lost on previous tasks. Given the interest in this phenomena, it is useful to explore if it occurs in the chunking setting and so see if it is only because of task shift or not.
\begin{figure}[t]
    \centering
    \subfloat{{\includegraphics[width=.5\linewidth]{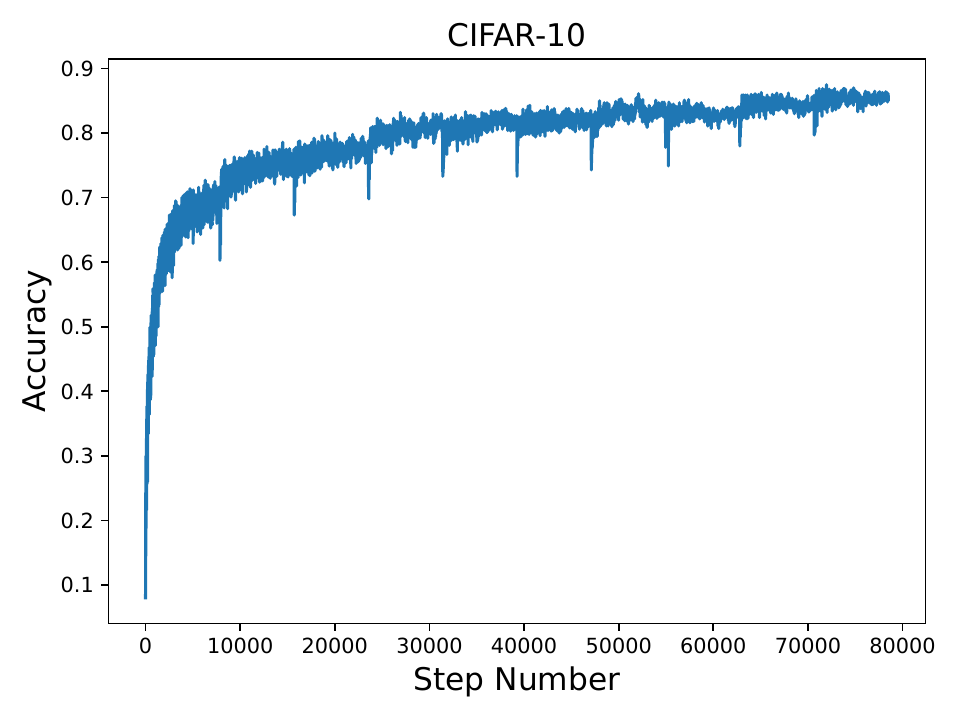} }}%
    \subfloat{{\includegraphics[width=.5\linewidth]{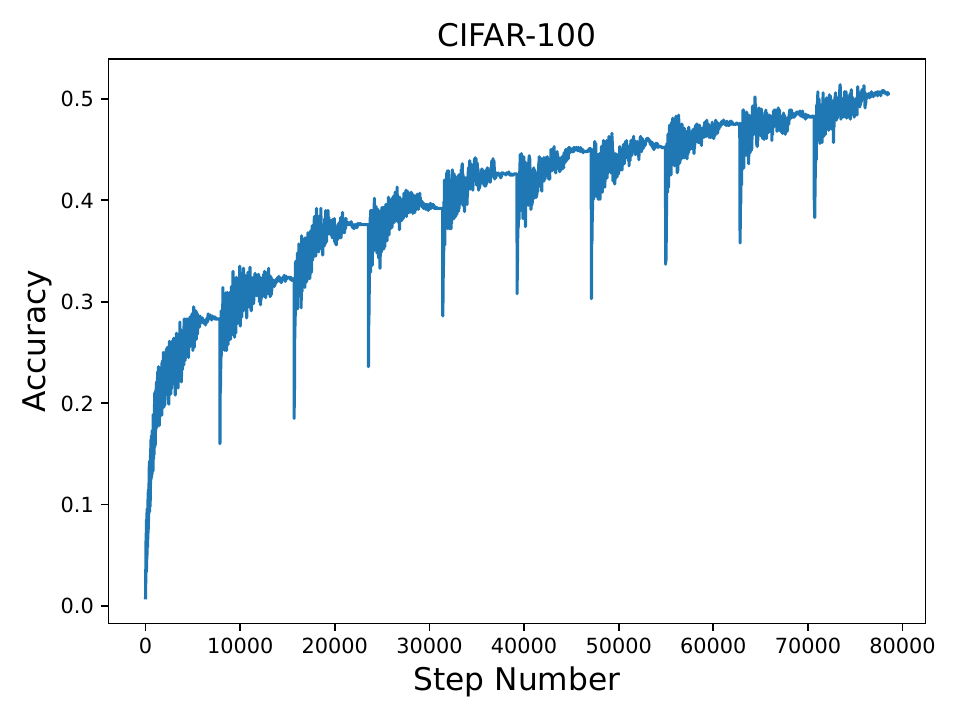} }}%
    \qquad
    \subfloat{{\includegraphics[width=.5\linewidth]{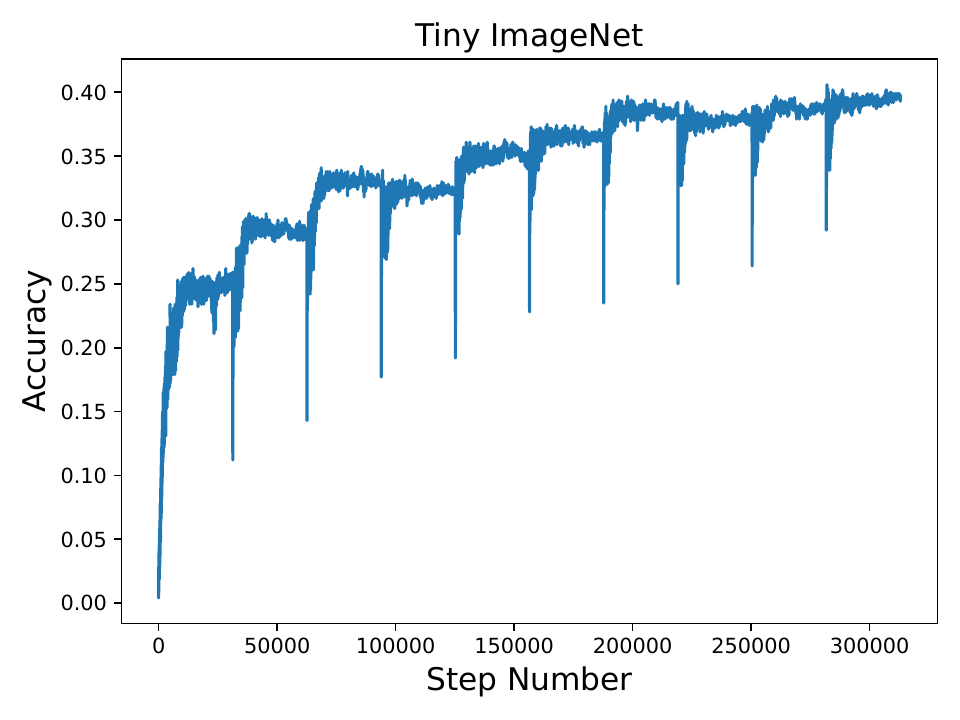} }}%
    \caption{Test accuracy curves for ER using 10 chunks on CIFAR-10, CIFAR-100 and Tiny ImageNet. Step number refers to the number of parameter updates which have been performed up to that point. The plots show that at the start of each chunk there is a large drop in test accuracy.}%
    \label{fig:ERStabilityGap}%
\end{figure}

We present experiments on the stability gap in the chunking setting in Figures~\ref{fig:SGDStabilityGap} and \ref{fig:ERStabilityGap}. The experiments consist of recording the performance on a held-out set of data for each parameter update step when learning on 10 chunks and using plain SGD training or ER. We use the same experimental setup as the rest of our experiments and these experiments are analogous to the ones performed by \citet{de2023continual} but for the chunking setting instead of standard CL. The results show that the stability gap also occurs in the chunking setting as all bar one of the plots show sudden drops in test performance when the learner starts learning on a new chunk. The plot which does not show sudden drops is  for plain SGD learning on CIFAR-10, where we believe the general noise in performance hides the systematic drops in performance at the start of each chunk. So, our results show that the stability gap phenomenon is not only to do with distribution shift. Furthermore, our experiments add light to why the stability gap occurs, where we can rule out distribution shift as being a necessary factor. Instead, these experiments indicate in conjunction with Figure~\ref{fig:50ChunkLossCurve}---which shows the training loss for plain SGD training---that there is some form of "overfitting" or compression \citep{tishby2015deep} occurring which induces high losses at the start of learning a chunk. This in turn leads to large update steps. Finally, these large updates potentially makes the network forget previously learnt information, creating the drop in performance at the start of a chunk we see in Figures~\ref{fig:SGDStabilityGap} and \ref{fig:ERStabilityGap}. Given that the stability gap occurs in the chunking setting a potential good direction to solve it would be to explore it in the setting. This is because, the chunking setting is simpler and easier to reason about than standard CL.

\FloatBarrier

\section{Analysis of the Linear Case}
\label{appen:LinCase}
To theoretically analyse the chunking problem we turn to the linear regression case, where we can leverage closed form solutions. In this case, the naive solution is to perform least squares on each arriving chunk. However, as the least squares problem is convex and so does not depend on the initialised weights, it will fully forget all the past chunks, only using the last chunk to create the predictor. This means that the standard least squares solution to linear regression fails in the chunking setting. Instead a better solution is to use Bayesian linear regression \citep{minka2000bayesian}. This is because Bayesian linear regression given any particular chunking of the data will return the same predictor and so fully solves the chunking setting. Therefore, it is instructive to see how Bayesian linear regression prevents forgetting. To achieve this we present below the update equations for Bayesian linear regression. The prior on the weights is $\thetaB \sim \mathcal{N}(\mathbf{0}, \VB_{0})$ and the posterior after seeing all the chunks up to and including the $(k-1)$th is $\thetaB | C_{1:k-1} \sim \mathcal{N}(\mB_{k-1}, \VB_{k-1})$. Additionally, for a chunk $C_t$ we define $\XB_t$ as its row-wise matrix of data instances and $\yB_t$ as its vector of targets. The likelihood is defined by assuming $y|\xB, \thetaB \sim \mathcal{N}(\thetaB^T\xB, \sigma^2)$. Then, the Bayesian posterior for the $k$th chunk is
\begin{align}
    \thetaB | C_{1:k} &\sim \mathcal{N}(\mB_{k}, \VB_{k}), \\
    \mB_{k} &= \VB_{k}\VB^{-1}_{k-1}\mB_{k-1}+\frac{1}{\sigma^2}\VB_{k}\XB_{k}^{T}\yB_{k}, \\
    \VB_{k}^{-1} &= \VB_{k}^{-1} + \frac{1}{\sigma^2}\XB^{T}_{k}\XB_{k}.
\end{align}
By recursively expanding the $\mB_{k-1}$ and $\VB_{k-1}$ terms till we reach the prior we have that
\begin{align} \label{eq: BLR}
    \mB_{k} &= \frac{1}{\sigma^2}\sum_{t=1}^{k} \VB_{k}\XB^{T}_{t}\yB_{t}, \\
    \VB_{k}^{-1} &= \VB_{0}^{-1}+\frac{1}{\sigma^2}\sum_{t=1}^{k} \XB^{T}_{t}\XB_{t} = \VB_{0}^{-1}+\frac{1}{\sigma^2}\XB^{T}_{1:k}\XB_{1:k}.
\end{align}

The equations above show that Bayesian linear regression prevents forgetting by having its posterior mean $\mB_{k}$ be: (a) a sum of the least squares solutions of each chunk and (b) instead of using the chunks unnormalised empirical covariance $\XB^{T}_{t}\XB_{t}$ in the least squares solutions it uses the running estimate of the weight precision $\VB_{k}^{-1}$. Computing and storing $\VB_{k}^{-1}$ is infeasibly costly for very large systems (e.g. neural networks), taking up $O(\text{dim}(\thetaB)^2)$ space. Therefore, assuming there is only enough memory to store a set of weights a backoff is to use a sum of the least squares solutions to each chunk. This is achieved by \emph{weight averaging}, where at each chunk we perform least squares on that chunk and add it to a running average, which results in the update equation,
\begin{align}
    \mB_{k} = \frac{k-1}{k}\mB_{k-1}+\frac{1}{k}(\XB^{T}_{k}\XB_{k})^{-1}\XB_{k}^{T}\yB_{k}. 
\end{align}
Again, by recursively expanding $\mB_{k-1}$ we have that,
\begin{align} \label{eq: WA}
    \mB_{k} = \frac{1}{k}\sum_{t=1}^{k} (\XB^{T}_{t}\XB_{t})^{-1}\XB^{T}_{t}\yB_{t}. 
\end{align}
Weight averaging gives similar, mean, weights as Bayesian linear regression where instead of using $\VB_{k}$ it uses the per-chunk estimate $\frac{1}{k}(\XB^{T}_{t}\XB_{t})^{-1}$ and we divide by $k$ to correctly scale the estimate. Both $\VB_{k}$ and $\frac{1}{k}(\XB^{T}_{t}\XB_{t})^{-1}$ are unnormalised estimates of the precision of the data distribution. Therefore, when each chunk is large enough that they are both accurate estimates, we have that $\frac{1}{k}(\XB^{T}_{t}\XB_{t})^{-1} \approx \VB_{k}$ for all $t\in \{1,\ldots,k\}$. In this case, weight averaging approximates Bayesian linear regression well and so should not forget that much. More formally, by using a concentration bound on the covariance estimates, we have the following theorem on the approximation error.\footnote{We focus on the goodness of the approximation in the size of each chunk and use simple techniques to give such a bound. With more complicated techniques, it should also be possible to examine the behaviour in the number of chunks, were we believe there should be some convergence to a fixed approximation error (a.s.).}
\begin{theorem}\label{thm:approx}
(proved in Appendix~\ref{appen:proof}) Assume that we have $k$ chunks and that each chunk $C_t = \{\xB_i \in \mathbb{R}^{d} | i = 1,...,S\}$ is sampled i.i.d. from an $\alpha$-sub-Gaussian distribution (assuming zero mean) with a full rank covariance matrix $\SigmaB$. Also, assume bounded random variables such that $\lVert\xB_i \rVert_2 \leq a_{\xB}$ and $\lVert \yB_t \rVert_2 \leq a_{\yB}$. Then, for the Bayesian linear regression model set the prior such that $\VB_0 = b \IB$ and let $b \rightarrow \infty$. Last, denote $\mB_{\textrm{BLR}}$ and $\mB_{\textrm{WA}}$ as the parameter estimates given by Bayesian linear regression and weight averaging, respectively. We then have for universal constants $a_1, a_2, a_3$ and for $\delta$ in the range $ \alpha^{-2} \lambda_d(\SigmaB) > \delta \geq 0$ the following approximation bound,
\[ \lVert \mB_{\textrm{BLR}} - \mB_{\textrm{WA}} \rVert_2 \leq \frac{2 a_{\xB} a_{\yB}}{\sqrt{S}} \frac{\epsilon(S, \delta)}{(\lambda_d(\SigmaB) - \epsilon(S, \delta))^2} \]
with probability of at least $1- k a_2 e^{- a_3 S \min(\delta, \delta^2)}$. Defining $\lambda_d(\SigmaB)$ as the smallest eigenvalue of $\SigmaB$,
\[ \epsilon(S, \delta) = \alpha^2 \left[a_1\left(\sqrt{\frac{d}{S}}+\frac{d}{S}\right)+\delta\right]\]
and assuming
\begin{align} \label{eq:s}
  S \geq \frac{\alpha^2 a_1 \left[\alpha^2 a_1 + 2(\lambda_d(\SigmaB) - \alpha^2 \delta) + \alpha\sqrt{a_1 (\alpha^2 a_1 +4(\lambda_d(\SigmaB) - \alpha^2 \delta))}\right]}{2(\lambda_d(\SigmaB) - \alpha^2 \delta)^2} d.
\end{align}
\end{theorem}
Theorem~\ref{thm:approx} shows formally that the larger the chunk size $S$ the better weight averaging approximates Bayesian linear regression and that in the limit weight averaging converges to Bayesian linear regression (a.s.).

The analysis in this section shows that weight averaging is a reasonable method for the linear case. It greatly improves performance over standard linear regression, which forgets all but the last chunk, and approximates well Bayesian linear regression which fully solves continual learning. Hence, the question arises if this analysis showing weight averaging improves performance also holds true for neural networks. We show in the main paper that empirically this is true (e.g. see Figure~\ref{fig:WeightAvgChunkingCurves}) but, to the best of our knowledge, it is an open problem to show it theoretically as well.

\section{The Effect of Pretraining in the Chunking Setting}
\label{appen:pretrain}
\begin{figure}[h]
    \centering
    \subfloat{\includegraphics[width=.5\linewidth]{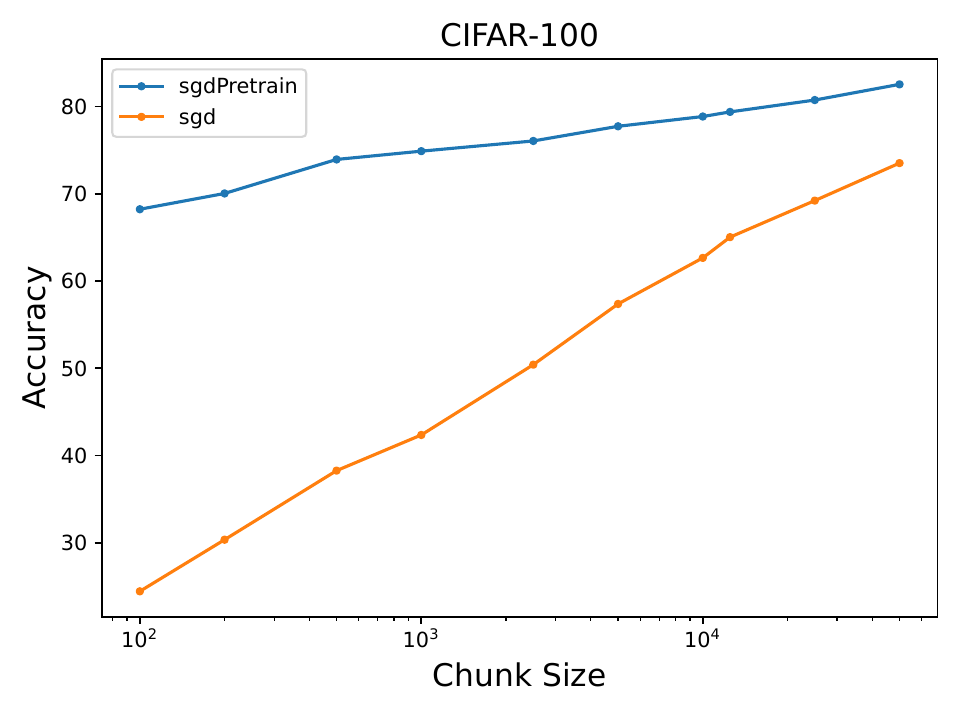}}
    \subfloat{{\includegraphics[width=.5\linewidth]{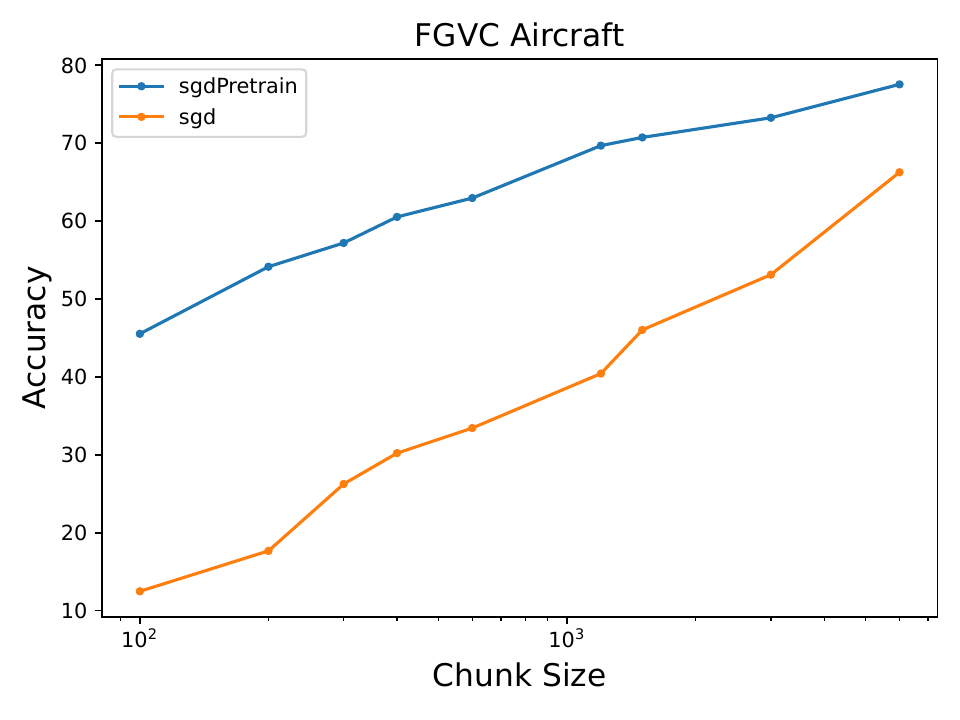} }}%
    \caption{End-of-training accuracy against chunk size on CIFAR-100 and FGVC Aircraft when finetuning a pretrained ResNet18 (sgdPretrain) or training from scratch (sgd), where the pretrained network is pretrained on ImageNet. Each data point on a curve presents the end-of-training accuracy of a method from a full run with chunks of the size given on the horizontal axis. The plots show that using a pretrained network reduces the performance gap from offline learning (the performance of the right most point for each curve) to chunking performance. But, for FGVC Aircraft which is more dissimilar to the pretraining dataset, the narrowing of the performance gap is less.} 
    \label{fig:PretrainChunkingCurve}
\end{figure}
While pretraining is often not considered in the CL literature it can greatly reduce the performance drop of CL methods compared to offline training \citep{hayes2020lifelong, ostapenko2022foundational, pelosin2022simpler}. Therefore, we present here how much pretraining aids in solving the chunking setting. In Figure~\ref{fig:PretrainChunkingCurve} we record the performance of using a ResNet18 model pretrained on ImageNet \citep{wightman2021resnet} compared from learning from scratch, using plain SGD training, for a dataset which is similar to ImageNet, CIFAR-100, and one which is not as similar, FGVC Aircraft \citep{maji2013fine}. The reason why we consider FGVC Aircraft to be not as similar to ImageNet as CIFAR-100 is that CIFAR-100 shares classes with ImageNet while FGVC Aircraft contains classes of different aircraft and so, given that ImageNet only has one aircraft class, FGVC Aircraft a is more fine-grained classification dataset. Figure~\ref{fig:PretrainChunkingCurve} shows that for CIFAR-100 pretraining greatly reduces the performance drop from offline learning to the chunking setting. For example, when using the pretrained network, there is only a difference of around $10\%$ between seeing the CIFAR-100 as one chunk (offline performance) and seeing chunks of 100 examples each. However, for FGVC Aircraft, while pretraining improves performance, there is still a significant performance gap to offline learning, around a $30\%$ accuracy drop, for small chunk sizes. Therefore, we have shown that pretraining helps solve the chunking problem but for data streams which are dissimilar to the pretraining dataset chunking is still a significant problem.

\section{Analysis of Stratified Sampling for Chunks}
\label{Appen:Sampling}
\begin{figure}[h]
    \centering
    \subfloat{{\includegraphics[width=.5\linewidth]{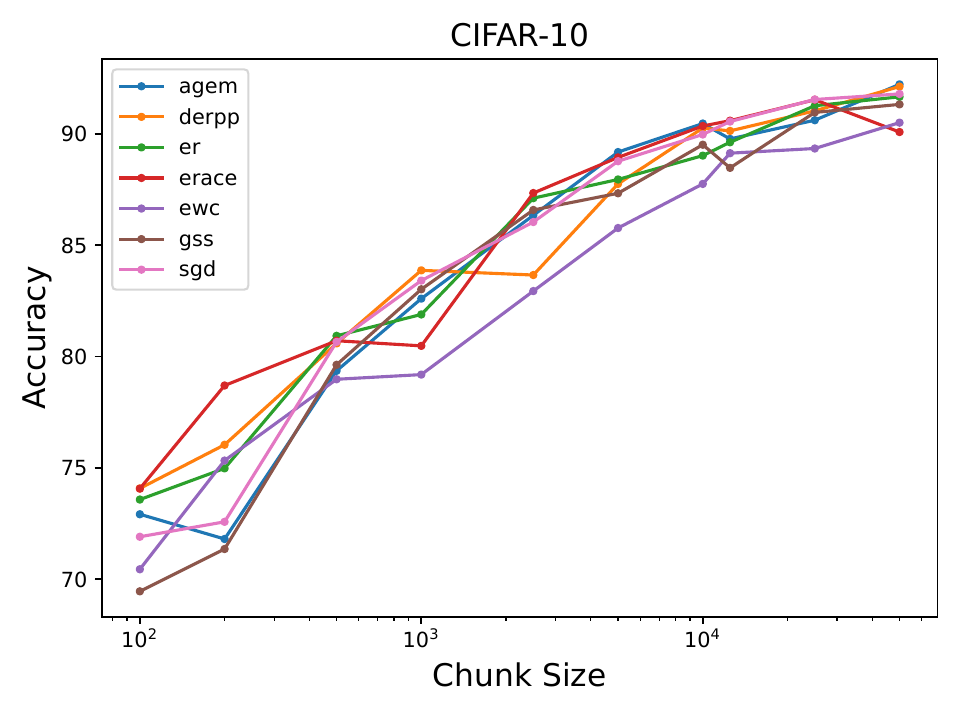} }}%
    \subfloat{{\includegraphics[width=.5\linewidth]{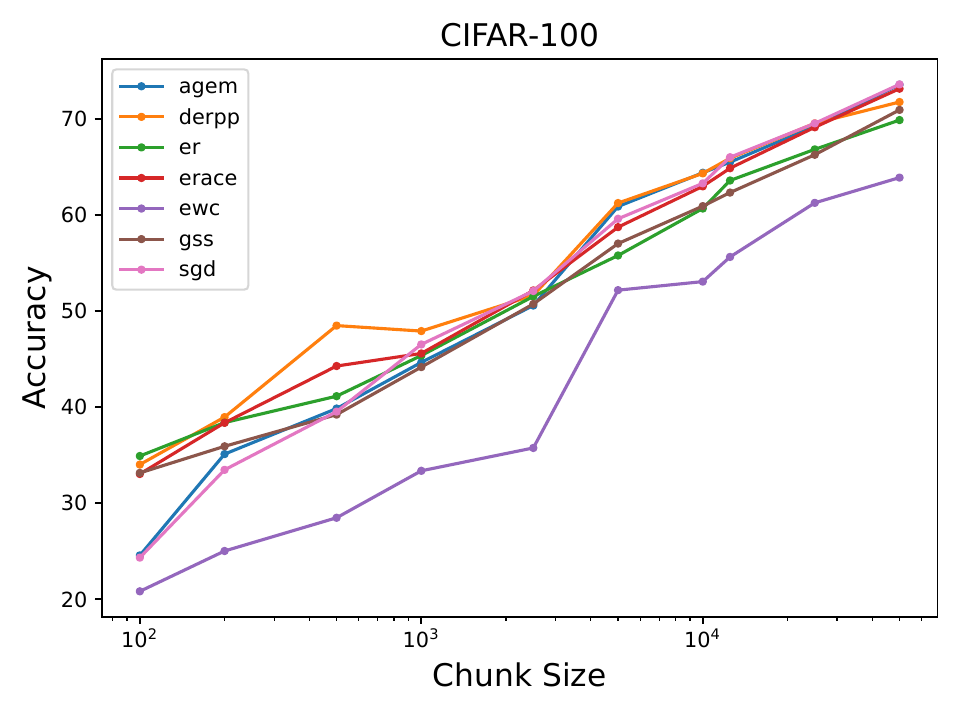} }}%
    \qquad
    \subfloat{{\includegraphics[width=.5\linewidth]{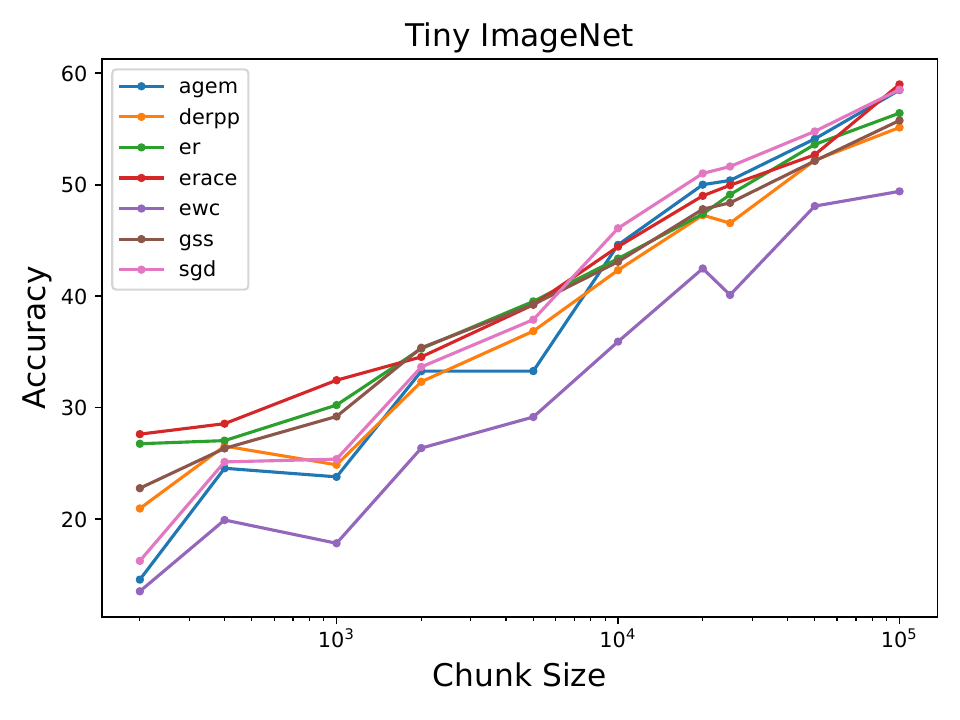} }}%
    \caption{Plots of end-of-training accuracy against chunk size on CIFAR-10, CIFAR-100 and Tiny ImageNet, where the data sampled for each chunk is not constrained to be class balanced. Each data point on a curve presents the end-of-training accuracy of a method from a full run with chunks of the size given on the horizontal axis. The plots show that sampling each chunk without ensuring they are class balanced gives the same trend as when the chunks are class balanced, which are displayed in Figures~\ref{fig:NNChunkingCurves} and \ref{fig:tinyimgChunkingCurve}. }%
    \label{fig:UnStratifiedNNChunkingCurves}%
\end{figure}
To make sure for small chunk sizes the drop in performance is due to the online availability of data and not class imbalance in a chunk, in the (balanced) chunking setting we stratify sample chunks such that each chunk has an equal amount of data from each class. However, in Figure~\ref{fig:UnStratifiedNNChunkingCurves} we look at what happens if the chunks were sampled randomly without ensuring the classes are balanced in each chunk (i.e., we randomly split the dataset into the given number of chunks). The figure shows the same trend as the when the chunks are classed balanced, displayed in Figures~\ref{fig:NNChunkingCurves} and \ref{fig:tinyimgChunkingCurve}, indicating that class imbalance does not affect our findings. However, class imbalance could be a problem for other datasets/objectives and is for the linear case therefore we assume the chunks are class balanced in the rest of this work and in our formulation of the chunking setting.  

\section{Analysis of Number of Epochs per Chunk}
\label{appen:epochNum}

\FloatBarrier

\begin{figure}[h]
    \centering
    \subfloat{{\includegraphics[width=.5\linewidth]{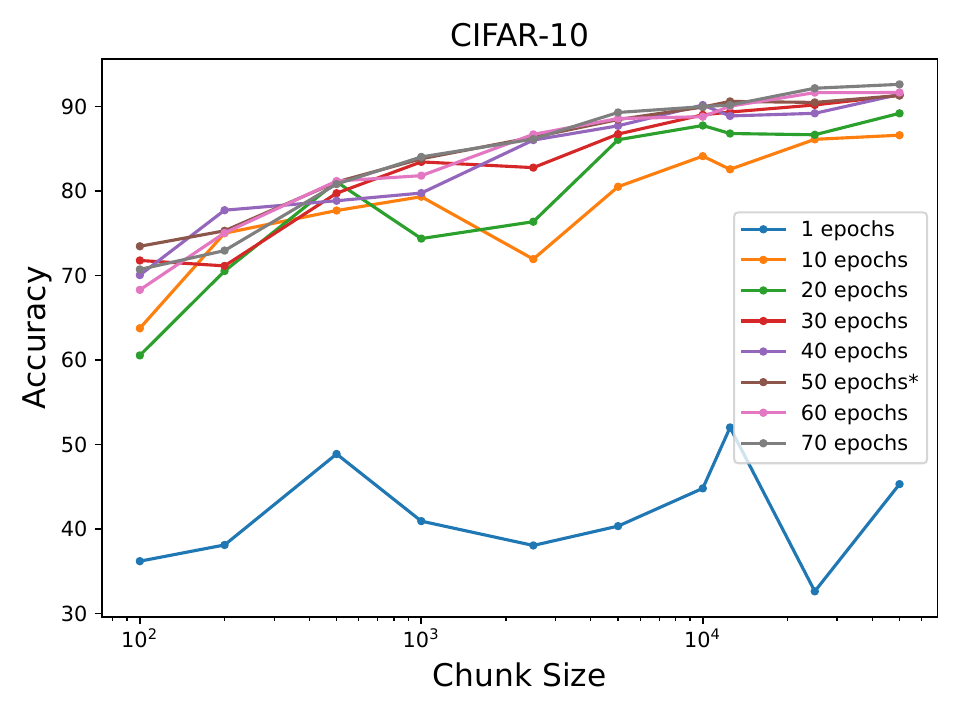} }}
    \subfloat{{\includegraphics[width=0.5\linewidth]{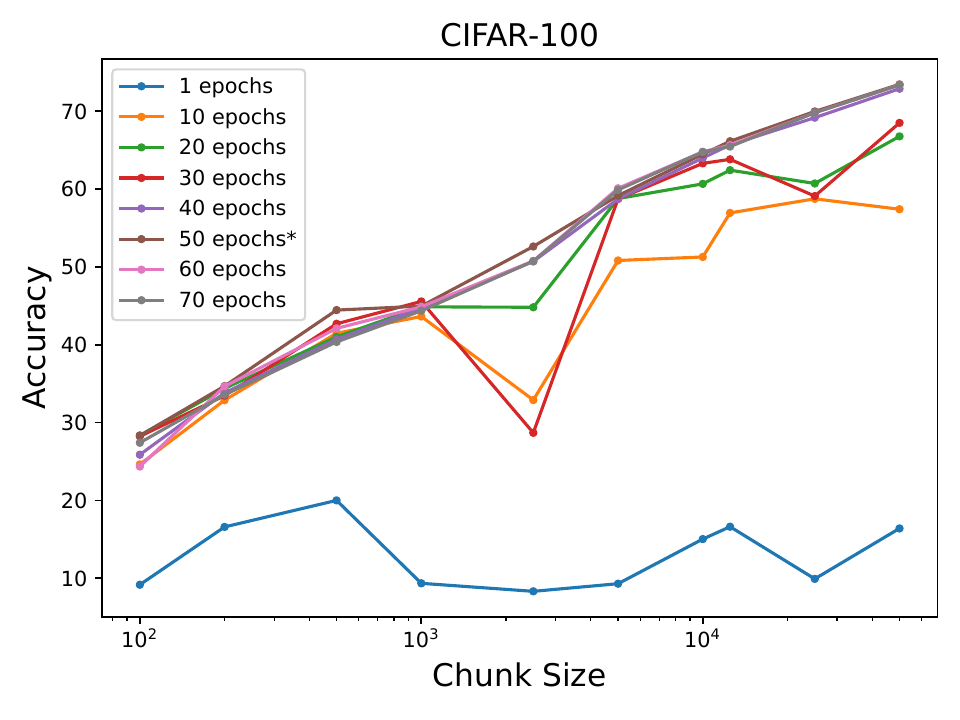}}}
    \qquad
    \subfloat{{\includegraphics[width=.5\linewidth]{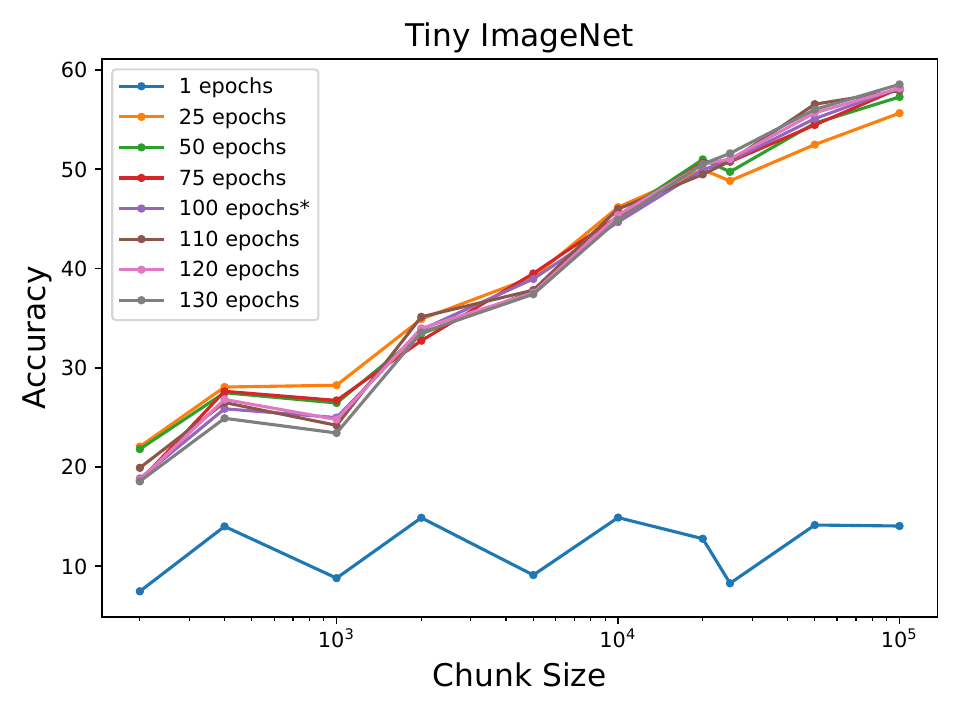} }}%
    \caption{Plots of end-of-training accuracy against chunk size on CIFAR-10, CIFAR-100 and Tiny ImageNet for SGD using different number of epochs per chunk. Each data point on a curve presents the end-of-training accuracy of a method from a full run with chunks of the size given on the horizontal axis. The `$^*$' denotes the number of epochs we use in the rest of the experiments and the figure shows that using our selected values achieves the best, or very similar to the best, accuracy for each chunk size. }
    \label{fig:n_epochsCurves}
\end{figure}
Hyperparameters can affect the reason why a method performs badly in the chunking setting. For example, if a method trains for one epoch over each chunk then it probably has bad performance due to underfiting; while, if it is trained for a reasonable number of epochs the reason would be forgetting, as shown in the main paper (e.g. Figure~\ref{fig:50ChunkForgettingCurve}). To remove this dependence we report the behaviour when using the hyperparameters which achieve the best accuracy for each chunk size. In Figure~\ref{fig:n_epochsCurves} we report the effect the number of training epochs used for each chunk has on performance, showing our selected number of epochs is the best or comparable to the best for each chunk size. Interestingly, the best performing number-of-epochs is the same for the full CL setting with task shift \citep{buzzega2020dark, boschini2022class}. This suggests that when fitting the training hyperparameters in CL, we are in part implicitly fitting them to minimise the effect of chunking. 

\vspace{2cm}
\section{Additional Forgetting Curves}
\label{appen:add50ChunkForgettingCurve}
\begin{figure}[h]
    \centering
    \subfloat{{\includegraphics[width=.5\linewidth]{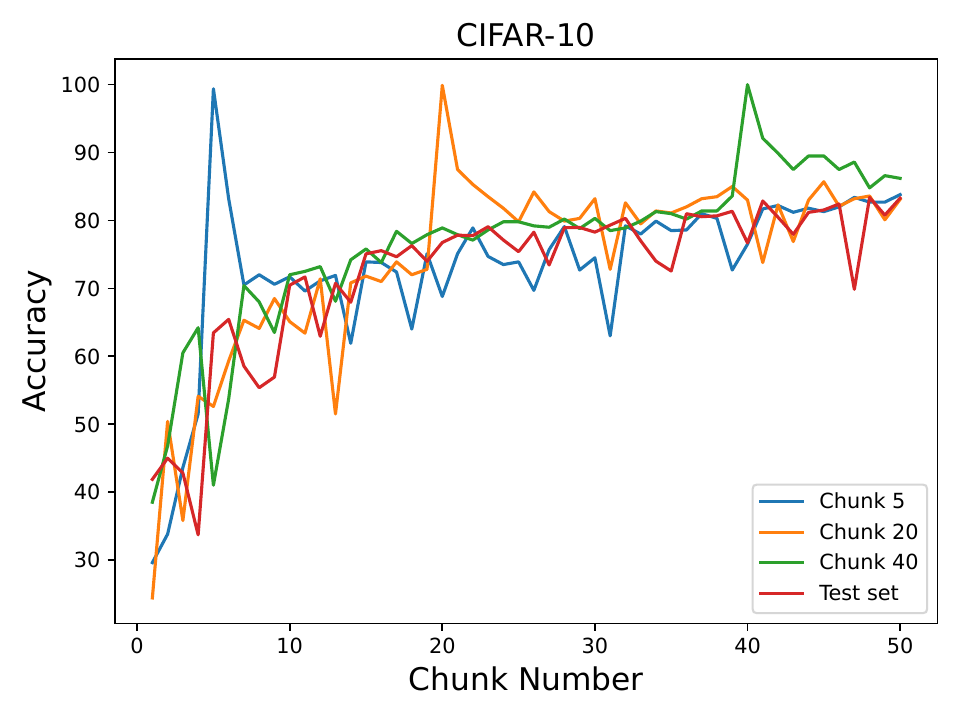} }}%
    \caption{Plot for plain SGD training of accuracy at the end of learning on each chunk for the training set of the $5^{th}$, $20^{th}$ and $40^{th}$ chunks and the test set, when training on CIFAR-10 with 50 chunks, corresponding to a chunk size of 1000. The plot shows that after learning on a chunk the accuracy on that chunk quickly drops to the level of test set performance and hence that the learner quickly forgets a large part of the knowledge of the chunk after learning on it.}%
    \label{fig:add50ChunkForgettingCurve}%
\end{figure}
\begin{figure}[h]
    \centering
    \subfloat{\includegraphics[width=.5\linewidth]{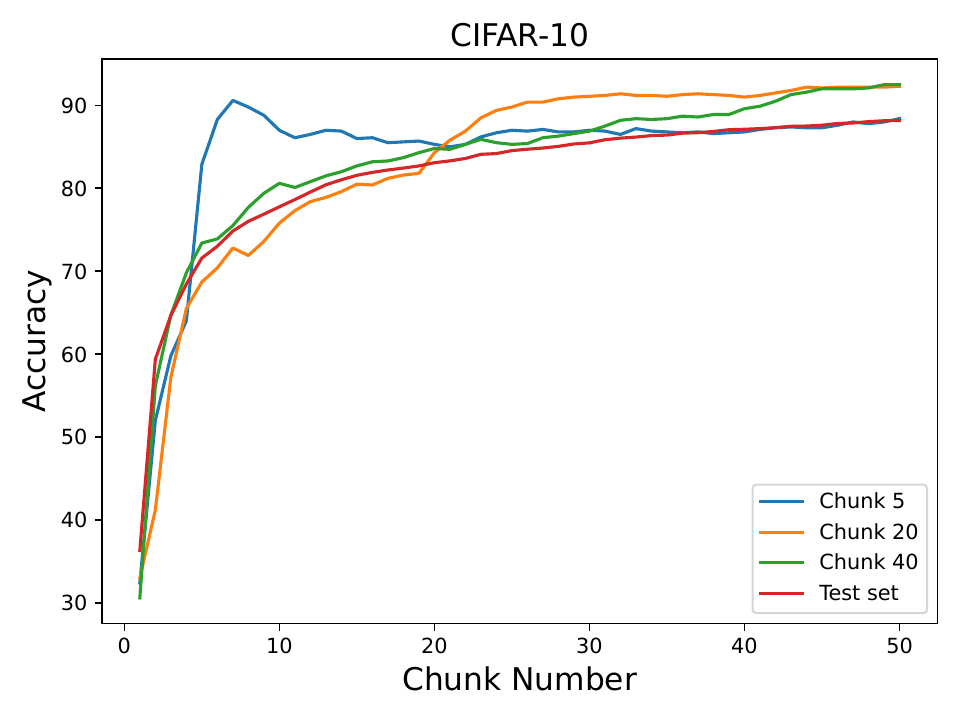}}
    \subfloat{{\includegraphics[width=.5\linewidth]{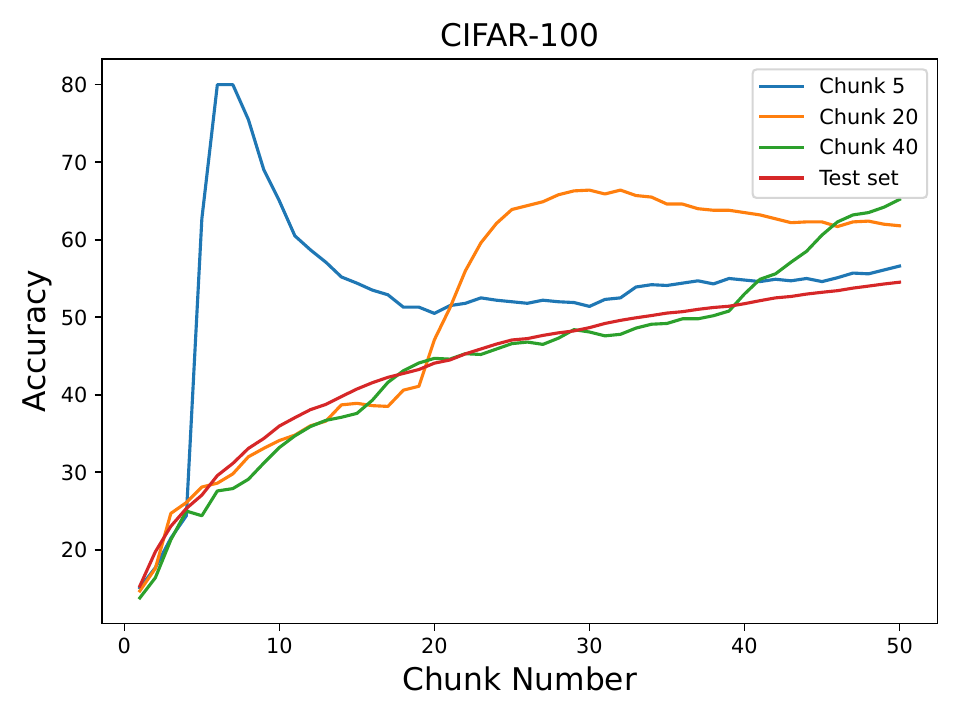} }}%
    \caption{Plots showing when using mean weight averaging the accuracy at the end of learning on each chunk for the training set of the $5^{th}$, $20^{th}$ and $40^{th}$ chunks and the test set, when training on CIFAR-10 and CIFAR-100 with 50 chunks, corresponding to a chunk size of 1000 for both datasets. The plots demonstrate that mean weight averaging forgets less than plain SGD training, whose analogous plots are displayed in Figures~\ref{fig:50ChunkForgettingCurve} and \ref{fig:add50ChunkForgettingCurve}.}
    \label{fig:addWeightAvgForgettingCurve}
\end{figure}

\FloatBarrier
\newpage
\section{Experiment on different weightings for EMA}
\label{appen:EMAChunkingCurves}
\begin{figure}[h]
    \vspace{-5mm}
    \centering
    \subfloat{{\includegraphics[width=.5\linewidth]{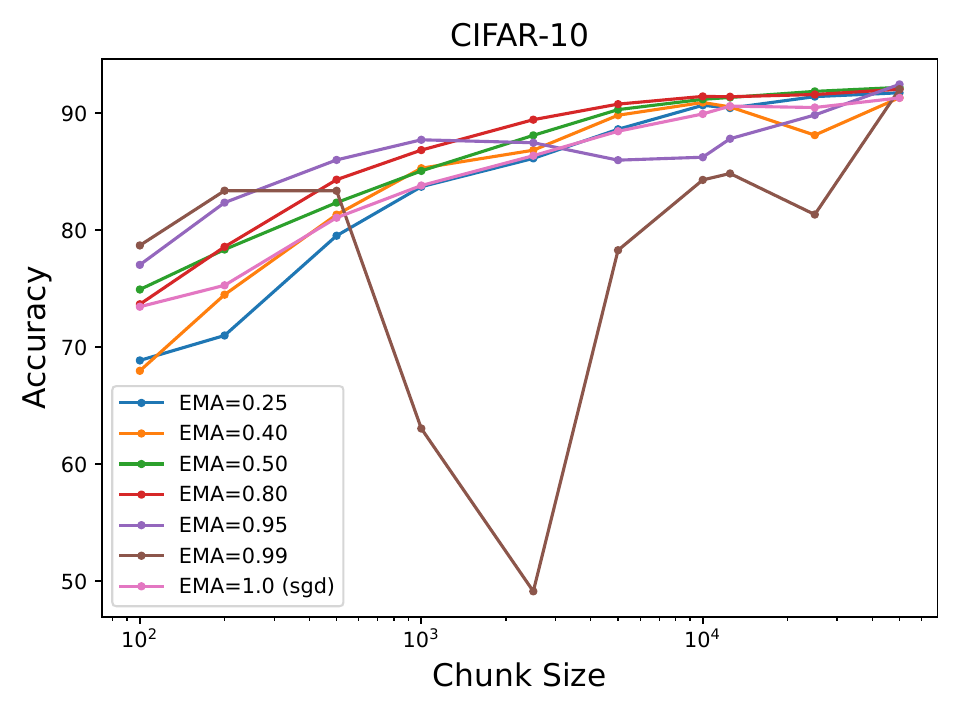} }}%
    \subfloat{{\includegraphics[width=.5\linewidth]{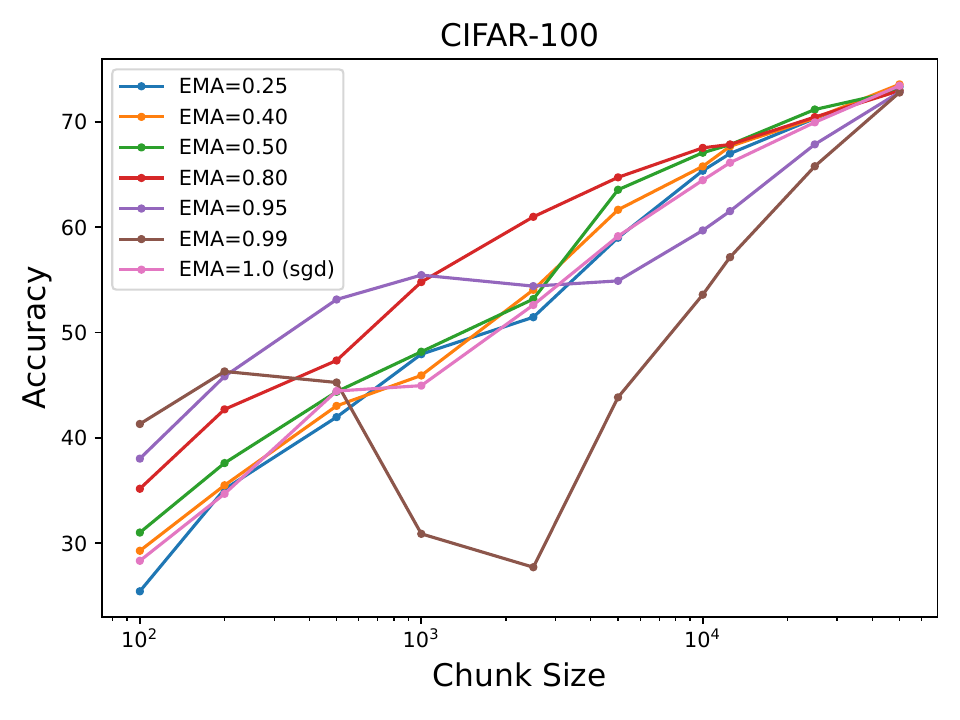} }}%
    \qquad
    \subfloat{{\includegraphics[width=.5\linewidth]{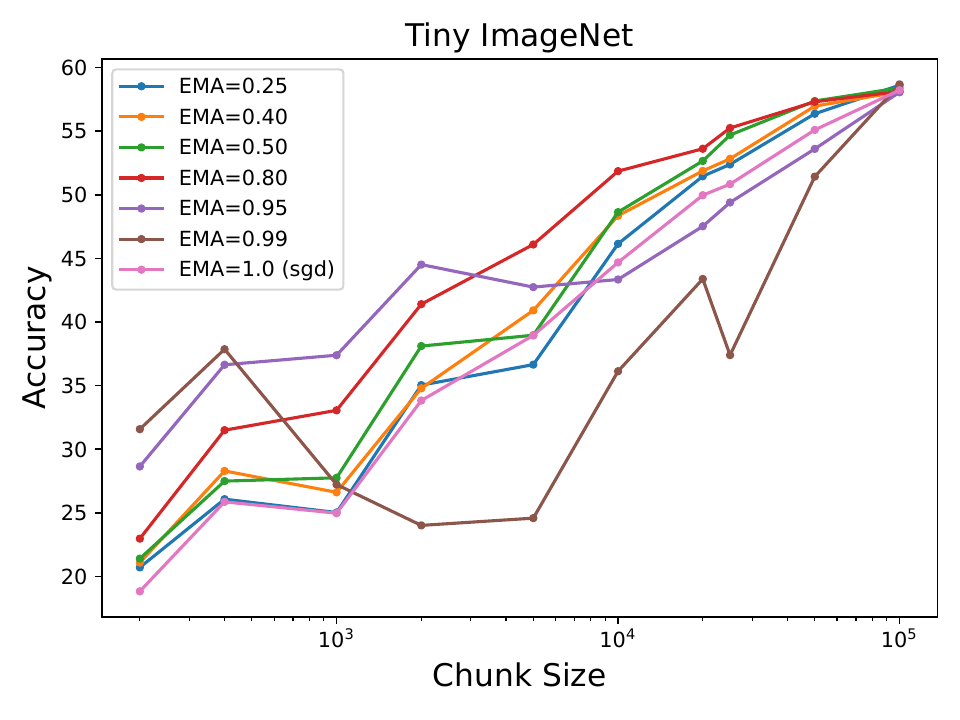} }}%
    \caption{Plots of the end-of-training accuracy when leaning with the given chunk size for different EMA weight values when learning with SGD on CIFAR-10, CIFAR-100 and tiny ImageNet. The plots shows that the EMA values of $0.80$ and $0.95$ achieve consistently the best or comparable to the best accuracy for all chunk sizes and datasets.}%
    \label{fig:EMAChunkingCurves}%
\end{figure}

\FloatBarrier
\section{Experiment on the performance of per-chunk EMA in standard CL}
\label{appen:EMACL}
\begingroup
\setlength{\tabcolsep}{2.5pt} 
\begin{table*}[h]
  \caption{Accuracy of ER in online and standard CL settings when using per-chunk mean averaging (WA-), per-chunk EMA (EMA-) or neither. The accuracies are averaged over 3 runs and where we report the standard error over the runs. The results show that using per-chunk mean averaging performs better than using per-chunk EMA.}
  \label{tab:EMACL}
  \centering
  \begin{tabular}{llllllll}
    \toprule
    & & \multicolumn{2}{c}{CIFAR-10} & \multicolumn{2}{c}{CIFAR-100} & \multicolumn{2}{c}{Tiny ImageNet} \\
    \cmidrule(r){3-4} \cmidrule(r){5-6} \cmidrule(r){7-8}
    Setting & Method & Class-IL & Task-IL  & Class-IL & Task-IL & Class-IL & Task-IL \\
    \midrule
    Online & ER & $36.19_{\pm 1.19}$ & $81.89_{\pm 0.92}$  & $8.45_{\pm 0.45}$ & $44.14_{\pm 1.31}$ & $5.56_{\pm 0.21}$  & $27.23_{\pm 0.65}$  \\
    & WA-ER & $39.59_{\pm 0.60}$ & $84.27_{\pm 0.37}$ & $14.01_{\pm 0.23}$ & $50.66_{\pm 0.77}$ & $7.77_{\pm 0.09}$ & $34.26_{\pm 0.33}$ \\
    & EMA-ER & $31.92_{\pm 0.45}$ & $79.22_{\pm 0.94}$ & $8.65_{\pm 0.66}$ & $41.72_{\pm 0.88}$ & $5.8_{\pm 0.13}$ & $28.24_{\pm 0.28}$  \\
    \midrule
    Standard & ER & $40.01_{\pm0.81}$ & $89.79_{\pm 0.75}$  & $11.78_{\pm0.34}$ & $57.80_{\pm1.02}$ & $8.36_{\pm0.16}$  & $31.72_{\pm0.46}$  \\
    & WA-ER & $56.49_{\pm 0.87}$ & $94.28_{\pm 0.17}$ & $24.24_{\pm 0.64}$ & $70.07_{\pm 0.29}$ & $12.31_{\pm 0.19}$ & $46.71_{\pm 0.33}$  \\
    & EMA-ER & $38.84_{\pm 0.25}$ & $89.82_{\pm 0.18}$ & $11.68_{\pm 0.08}$ & $58.16_{\pm 0.90}$ & $8.20_{\pm 0.14}$ & $33.76_{\pm 1.72}$ \\
    \bottomrule
  \end{tabular}
\end{table*}
\endgroup
In the main paper we only look at per-chunk mean weight averaging in the CL experiments, here we also report results using per-chunk EMA weight averaging (Table~\ref{tab:EMACL}). The reason we only looked at per-chunk mean weight averaging in the main paper was that we showed it had better performance in the chunking setting. The results in Table~\ref{tab:EMACL} shows that in both online and standard CL it is still the case that per-chunk mean weight averaging performs better. This shows a correspondence between the performances in the chunking and CL settings. One point to note is that we use weighting of 0.8 for EMA in these experiments.  

\section{Proof of Theorem~\ref{thm:approx}} \label{appen:proof}
In this section we prove a bound on how well weight averaging approximates Bayesian linear regression. To do this we first state the concentration bound used in the proof, where $||.||_2$ for a matrix denotes the $l_2$-operator norm.
\begin{theorem}[rephrased from \citet{wainwright2019high} (Theorem~6.5 p.166)] \label{thm:con}
Given a chunk $C_t = \{\xB_i \in \mathbb{R}^{d} | i = 1,...,S\}$ sampled i.i.d. from a $\alpha$-sub-Gaussian distribution (assuming zero mean) with covariance matrix $\SigmaB$; there are universal constants $a_1, a_2, a_3$ such that the empirical covariance matrix $\frac{1}{S} \XB_{t}^{T}\XB_t$ satisfies the bound,
\[ P\left(\frac{1}{\alpha^2} \left\lVert \frac{1}{S} \XB_{t}^{T}\XB_t - \SigmaB \right\rVert_2 \leq a_1\left[\sqrt{\frac{d}{S}}+\frac{d}{S}\right]+\delta\right) \geq 1 - a_2 e^{-a_3 S \min(\delta, \delta^2)} \] for all $\delta \geq 0$.
\end{theorem}

To allow us to complete the proof we need some assumptions which are given here. We assume that we see $k$ chunks and that each chunk $C_t = \{\xB_{t, i} \in \mathbb{R}^{d} | i = 1,...,S\}$ is sampled i.i.d. from an $\alpha$-sub-Gaussian distribution (assuming zero mean) with a full rank covariance matrix $\SigmaB$. We also assume bounded random variables such that $\lVert \xB_i \rVert_2 \leq a_{\xB}$ and $\lVert \yB_t \rVert_2 \leq a_{\yB}$. Additionally, for the Bayesian linear regression model we use a prior such that $\VB_0 = b \IB$ and let $b \rightarrow \infty$. To make the proof easier to follow we denote the running and per-chunk covariances as $\hat{\SigmaB}_{1:k} = \frac{1}{N S} \XB^{T}_{1:k} \XB_{1:k}$ and $\hat{\SigmaB}_{t} = \frac{1}{S} \XB^{T}_{t} \XB_{t}$, respectively. Furthermore, we denote the projected targets as $\yB'_{t} = \XB_{t}^{T} \yB_t$. Also, let $\mB_{\textrm{BLR}}$ and $\mB_{\textrm{WA}}$ be the parameter estimates given by Bayesian linear regression and weight averaging, respectively, and note to calculate such estimates we assume that the empirical precision matrices exists. In the proof we need a bound on the $\yB'_{t}$ for each chunk which is,
\begin{align} \label{eq:y}
    \lVert \yB'_{t} \rVert_2 &= \lVert \XB^{T}_{t} \yB_{t} \rVert_2 \\
    &\leq \lVert \yB_t \rVert_2 \max_{\vB, \lVert \vB \rVert_2 = 1}(\lVert \XB^{T}_{t} \vB \rVert_2) \\
    &= \lVert \yB_t \rVert_2 \lVert \XB^{T}_{t} \rVert_2 \\
    &\leq \lVert \yB_t \rVert_2 \lVert \XB^{T}_{t} \rVert_{F}\\
    &\leq  a_{\xB} a_{\yB} \sqrt{S}.
\end{align}
Additionally, we need to be able to bound $\lVert \hat{\SigmaB}_{1:k}^{-1} \rVert_2$ and $\lVert \hat{\SigmaB}_{t}^{-1} \rVert_2$. This is achieved using Weyl's inequality, where we have for $\lambda_d(\hat{\SigmaB}_{t})$ and $\lambda_d(\SigmaB)$,  the smallest eigenvalues for $\hat{\SigmaB}_{t}$ and $\SigmaB$, respectively, that 
\begin{align} \label{eq:weyl}
    \lambda_d(\hat{\SigmaB}_{t}) \geq  \lambda_d(\SigmaB) - \lVert \hat{\SigmaB}_{t} -\SigmaB  \rVert_2.
\end{align}
Now, if we were to use Theorem~\ref{thm:con} we can ensure the r.h.s. of Eq.~\ref{eq:weyl} is greater or equal to zero, by using a large enough chunk size $S$. This will be done in the proof and therefore we will have the following bound on $\lVert \hat{\SigmaB}_{t}^{-1} \rVert_2$,
\begin{align} \label{eq:norm}
    \lVert \hat{\SigmaB}_{t}^{-1} \rVert_2 = \frac{1}{\lambda_d(\hat{\SigmaB}_{t})} \leq \frac{1}{\lambda_d(\SigmaB) - \lVert \hat{\SigmaB}_{t} -\SigmaB  \rVert_2}.
\end{align}
We have an equivalent bound for $\lVert \hat{\SigmaB}_{1:k}^{-1} \rVert_2$ using the same argument and using the fact that $\hat{\SigmaB}_{1:k}$ is the mean of the chunk covariances $\hat{\SigmaB}_{t}$.

Given the above ground work, the proof proceeds as follows,
\begin{align}
    \lVert \mB_{\textrm{BLR}} - \mB_{\textrm{WA}} \rVert_2 &= \left\lVert \sum_{t=0}^{k} \frac{1}{k S} \hat{\SigmaB}_{1:k}^{-1} \yB'_{t} - \sum_{t=0}^{k} \frac{1}{k S} \hat{\SigmaB}_{t}^{-1} \yB'_{t} \right\rVert_2 \text{ \: \: \: (by simplifying from Eqs.~\ref{eq: WA} and \ref{eq: BLR})} \\
    &\leq \frac{1}{k S} \sum_{t=0}^{k} \lVert  \hat{\SigmaB}_{1:k}^{-1} \yB'_{t} - \hat{\SigmaB}_{t}^{-1} \yB'_{t} \rVert_2 \\
    &= \frac{1}{k S} \sum_{t=0}^{k} \lVert  [\hat{\SigmaB}_{1:k}^{-1} - \hat{\SigmaB}_{t}^{-1}] \yB'_{t} \rVert_2 \\
    &\leq \frac{1}{k S} \sum_{t=0}^{k} \lVert \yB'_{t} \rVert_2  \max_{\vB, \lVert \vB \rVert_2 = 1}( \lVert [\hat{\SigmaB}_{1:k}^{-1} - \hat{\SigmaB}_{t}^{-1}] \vB \rVert_2) \\
    &= \frac{1}{k S} \sum_{t=0}^{k} \lVert \yB'_{t} \rVert_2  \lVert \hat{\SigmaB}_{1:k}^{-1} - \hat{\SigmaB}_{t}^{-1} \rVert_2 \text{ \: \: \: (by variational def. of the operator norm)} \\
    &= \frac{1}{k S} \sum_{t=0}^{k} \lVert \yB'_{t} \rVert_2  \lVert  \hat{\SigmaB}_{1:k}^{-1} 
    (\hat{\SigmaB}_{1:k} - \hat{\SigmaB}_{t}) \hat{\SigmaB}_{t}^{-1}  \rVert_2 \\
    &\leq \frac{1}{k S} \sum_{t=0}^{k} \lVert \yB'_{t} \rVert_2  \lVert \hat{\SigmaB}_{1:k}^{-1} \rVert_2 \lVert \hat{\SigmaB}_{t}^{-1} \rVert_2 \lVert \hat{\SigmaB}_{1:k} - \hat{\SigmaB}_{t}  \rVert_2 \\
    &= \frac{1}{k S} \sum_{t=0}^{k} \lVert \yB'_{t} \rVert_2  \lVert \hat{\SigmaB}_{1:k}^{-1} \rVert_2 \lVert \hat{\SigmaB}_{t}^{-1} \rVert_2 \lVert (\hat{\SigmaB}_{1:k} -\SigmaB) - (\hat{\SigmaB}_{t} -\SigmaB)  \rVert_2 \\
    &\leq \frac{1}{k S} \sum_{t=0}^{k} \lVert \yB'_{t} \rVert_2  \lVert \hat{\SigmaB}_{1:k}^{-1} \rVert_2 \lVert \hat{\SigmaB}_{t}^{-1} \rVert_2 [\lVert \hat{\SigmaB}_{1:k} -\SigmaB \rVert_2 + \lVert \hat{\SigmaB}_{t} -\SigmaB  \rVert_2] \\
    &\leq \frac{a_{\xB} a_{\yB}}{k \sqrt{S} } \sum_{t=0}^{k} \lVert \hat{\SigmaB}_{1:k}^{-1} \rVert_2 \lVert \hat{\SigmaB}_{t}^{-1} \rVert_2 [\lVert \hat{\SigmaB}_{1:k} -\SigmaB \rVert_2 + \lVert \hat{\SigmaB}_{t} -\SigmaB  \rVert_2] \text{ \: \: \: (using Eq.~\ref{eq:y})} \\
    &\leq \frac{2 a_{\xB} a_{\yB}}{k \sqrt{S} } \sum_{t=0}^{k} \lVert \hat{\SigmaB}_{1:k}^{-1} \rVert_2 \lVert \hat{\SigmaB}_{t}^{-1} \rVert_2 \lVert \hat{\SigmaB}_{t} -\SigmaB  \rVert_2 \text{ \: \: \: (as } \hat{\SigmaB}_{1:k} \text{ is the mean of the chunk covs. } \hat{\SigmaB}_{t} \text{)}\\
\end{align}
Now by using Theorem~\ref{thm:con} to bound each $\lVert \hat{\SigmaB}_{t} -\SigmaB  \rVert_2$, the union bound, and Eq.~\ref{eq:norm} applied to $\lVert \hat{\SigmaB}_{1:k}^{-1} \rVert_2$ and each $\lVert \hat{\SigmaB}_{t}^{-1} \rVert_2$, we have that,
\begin{align}
    \lVert \mB_{\textrm{BLR}} - \mB_{\textrm{WA}} \rVert_2 \leq \frac{2 a_{\xB} a_{\yB}}{\sqrt{S}} \frac{\epsilon(S, \delta)}{(\lambda_d(\SigmaB) - \epsilon(S, \delta))^2}
\end{align}
for $\delta$ in the range $\alpha^{-2} \lambda_d(\SigmaB) > \delta \geq 0$ with probability of at least $1- k a_2 e^{- a_3 S \min(\delta, \delta^2)}$, where
\begin{align}
  \epsilon(S, \delta) = \alpha^2 \left[a_1\left(\sqrt{\frac{d}{S}}+\frac{d}{S}\right)+\delta\right]
\end{align}
and
\begin{align} \label{eq:s}
  S \geq \frac{\alpha^2 a_1 \left[\alpha^2 a_1 + 2(\lambda_d(\SigmaB) - \alpha^2 \delta) + \alpha\sqrt{a_1 (\alpha^2 a_1 +4(\lambda_d(\SigmaB) - \alpha^2 \delta))}\right]}{2(\lambda_d(\SigmaB) - \alpha^2 \delta)^2} d.
\end{align}
This completes the proof. We note that by using the union bound we remove the convergence behaviour in the number of chunks. A way to solve this problem would be to bound the mean over $\lVert \hat{\SigmaB}_{t}^{-1} \rVert_2 \lVert \hat{\SigmaB}_{t} -\SigmaB  \rVert_2$ using more complex methods.

\end{document}